\documentclass[sigconf]{acmart}

\usepackage{array}
\usepackage{subcaption}
\usepackage{multirow}
\usepackage{arydshln}
\usepackage{tabulary}
\usepackage{colortbl}
\usepackage{xcolor}
\usepackage{enumitem}
\usepackage{bm}
\usepackage{mathtools}
\usepackage{fixltx2e}
\usepackage{amsmath}
\usepackage{lipsum}
\usepackage{breqn} 
\usepackage{microtype} 
\usepackage[skip=4pt]{caption}
\usepackage{titlesec}
\usepackage{amssymb}
\titlespacing*{\section}{0pt}{10pt plus 4pt minus 2pt}{6pt plus 2pt minus 1pt}
\titlespacing*{\subsection}{0pt}{8pt plus 3pt minus 1pt}{4pt plus 1pt minus 1pt}
\titlespacing*{\subsubsection}{0pt}{6pt plus 2pt minus 1pt}{2pt plus 1pt minus 1pt}

\AtBeginDocument{%
  }

\setcopyright{acmlicensed}
\copyrightyear{2025}
\acmYear{2025}
\acmDOI{XXXXXXX.XXXXXXX}
\acmConference[KDD '25]{Proceedings of the 31st ACM SIGKDD Conference on Knowledge Discovery and Data Mining}{August 3--7, 2025}{Toronto, ON, Canada}

\acmISBN{978-x-xxxx-xxxx-x/YY/MM}

\acmSubmissionID{1900}

\begin{document}\sloppy

\title{DiSa: Directional Saliency-Aware Prompt Learning for Generalizable Vision-Language Models}

\author[Alipour Talemi et al.]{Niloufar Alipour Talemi}
\email{nalipou@clemson.edu}

\affiliation{%
  \institution{Clemson University}
  \city{Clemson}
  \state{SC}
  \country{United States}
}

\author[ ]{Hossein Kashiani}
\email{hkashia@clemson.edu}

\affiliation{%
  \institution{Clemson University}
  \city{Clemson}
  \state{SC}
  \country{United States}
}

\author[ ]{Hossein R. Nowdeh}
\email{hrajoli@clemson.edu}

\affiliation{%
  \institution{Clemson University}
  \city{Clemson}
  \state{SC}
  \country{United States}
}

\author[ ]{Fatemeh Afghah}
\email{fafghah@clemson.edu}

\affiliation{%
  \institution{Clemson University}
  \city{Clemson}
  \state{SC}
  \country{United States}
}

\begin{abstract}

Prompt learning has emerged as a powerful paradigm for adapting vision-language models such as CLIP to downstream tasks. However, existing methods often overfit to seen data, leading to significant performance degradation when generalizing to novel classes or unseen domains. To address this limitation, we propose DiSa, a Directional Saliency-Aware Prompt Learning framework that integrates two complementary regularization strategies to enhance generalization. First, our Cross-Interactive Regularization (CIR) fosters cross-modal alignment by enabling cooperative learning between prompted and frozen encoders. Within CIR, a saliency-aware masking strategy guides the image encoder to prioritize semantically critical image regions, reducing reliance on less informative patches. Second, we introduce a directional regularization strategy that aligns visual embeddings with class-wise prototype features in a directional manner to prioritize consistency in feature orientation over strict proximity. This approach ensures robust generalization by leveraging stable prototype directions derived from class-mean statistics. Extensive evaluations on 11 diverse image classification benchmarks demonstrate that DiSa consistently outperforms state-of-the-art prompt learning methods across various settings, including base-to-novel generalization, cross-dataset transfer, domain generalization, and few-shot learning.

\begin{figure}[t]

    \centering 
    \includegraphics[scale=0.31]{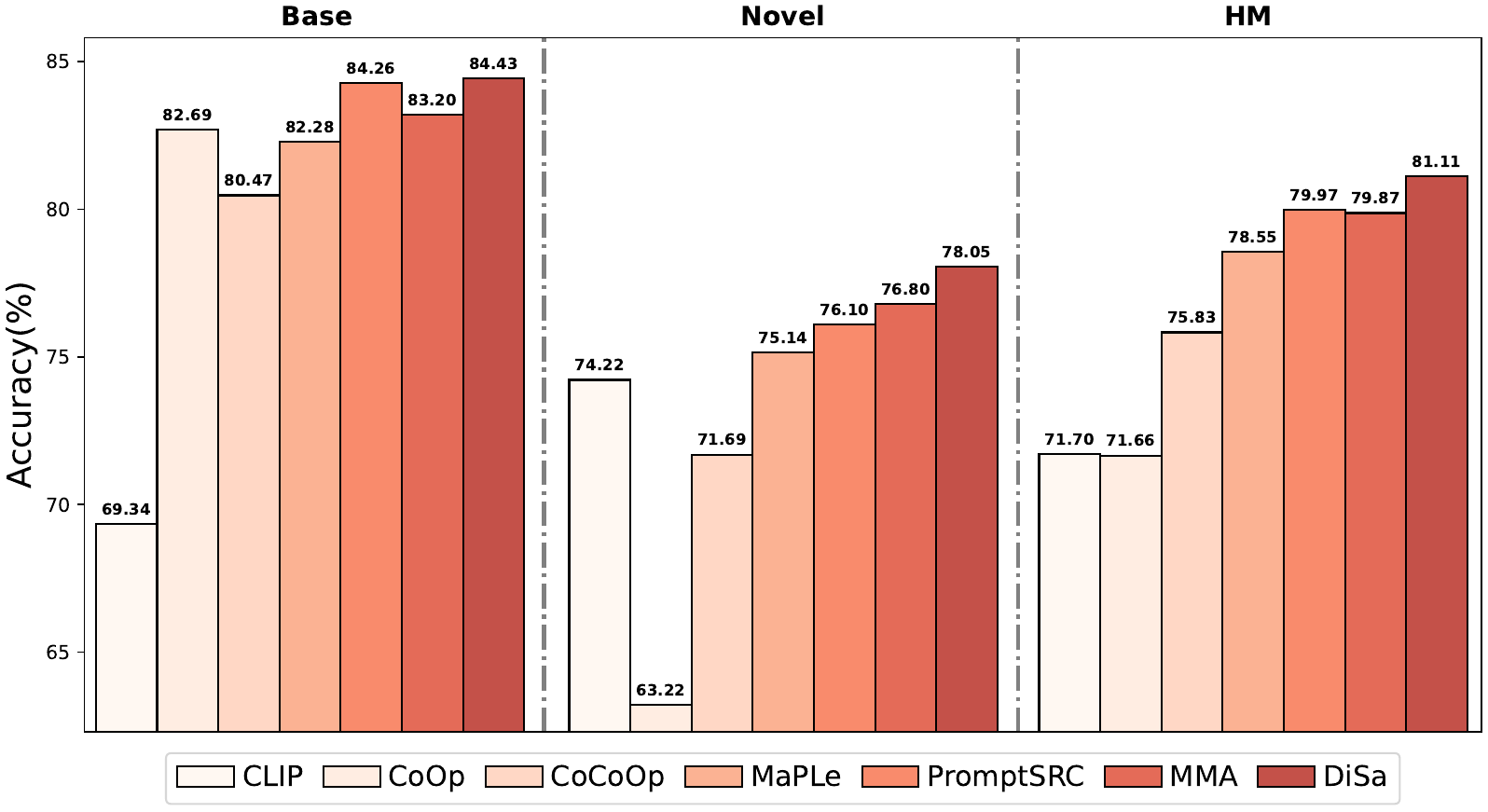}
    \caption{Performance comparison of base-to-novel generalization: DiSa outperforms state-of-the-art methods across 11 diverse image recognition datasets for both base and novel classes.}
    \label{fig:first}%
\end{figure}

\end{abstract}

\ccsdesc[500]{Computing methodologies~Computer vision}

\keywords{Vision-Language Models, Few-shot Learning, Prompt Learning, Domain Generalization}

\maketitle

\section{Introduction}
Large-scale pre-trained Vision-Language (VL) models have become foundational tools across a broad range of downstream tasks including applications such as few-shot image recognition \cite{kim2021adapt, gao2024clip}, object detection \cite{feng2022promptdet}, and image segmentation \cite{ding2022decoupling, he2023clip}. These models, trained on enormous web-scale datasets, show exceptional generalization ability across diverse downstream tasks without task-specific tuning. Among these models, Contrastive Language-Image Pre-training (CLIP) \cite{radford2021learning} stands out as an innovative model, leveraging contrastive learning on massive image-text pairs from the internet. While CLIP excels in zero-shot recognition, fine-tuning it for downstream tasks remains challenging, particularly in the few-shot setting, where limited data can cause overfitting and compromise its generalization ability \cite{zhou2022learning}.

The adaptation of CLIP to downstream recognition tasks has been explored through various methods, with prompt engineering \cite{radford2021learning} being a straightforward yet labor-intensive approach. It involves crafting input queries, such as ``a photo of a'' or ``a close-up photo of a'',  to condition the model’s text encoder for category-specific embeddings. However, this process demands extensive manual tuning and does not guarantee optimal prompts for target tasks \cite{zhou2022learning}. As a more efficient alternative, prompt learning has been introduced to adapt CLIP to downstream tasks without modifying its pre-trained weights \cite{zhou2022conditional, zhou2022learning, bulat2023lasp, rao2022denseclip, tsimpoukelli2021multimodal, khattak2023maple}. This method incorporates a small number of learnable prompt vectors, allowing the model to better align with downstream objectives. However, since these prompts are optimized based on task-specific objectives, the adapted model can overfit to the training distribution \cite{zhou2022conditional}. Consequently, while the model may achieve strong performance on seen classes, its performance on unseen classes or domains can degrade significantly.


Recently, some studies \cite{roy2023consistency, khattak2023self} have introduced regularization approaches to jointly optimize for both task-specific objectives and task-agnostic general representations to mitigate prompt overfitting. These approaches primarily focus on maximizing score similarity between the prompted and frozen models. However, they often overlook the significant contributions of individual modalities when leveraging their cross-modal alignment. These approaches treat vision and text modalities in isolation, neglecting their inherent cross-modal relationships. This limitation reduces the effectiveness of cross-modal learning, as the model cannot fully leverage the complementary strengths of modalities. We argue that effective regularization should also preserve cross-modal alignment. In this work, we propose a Cross-Interactive Regularization (CIR) framework that fosters a deeper interaction between the modality-specific branches of each model. This CIR framework enables cooperative learning by facilitating the interaction of image embeddings from the prompted model with text embeddings from the frozen model, and vice versa. This structure encourages the prompted encoders to independently engage with the cross-modal representations from the pre-trained encoders, leveraging the generalization capabilities of the pre-trained VL model while adapting prompts for downstream tasks. As such, the CIR framework prevents modality-specific overfitting by anchoring the prompted encoders to the frozen encoder’s cross-modal relationships.

Furthermore, training with few samples often results in the prompted image encoder developing a dependency on specific patches, which undermines its ability to generalize to unseen classes and data. To address this issue, we equip our CIR framework with a saliency-aware masking approach that guides the prompted image encoder to focus on critical patches under the guidance of the frozen model. More precisely, the masking process involves identifying less informative patches based on the guidance of the frozen model and then applying random masking within this subset of patches. This selective randomness ensures that the masked images retain the most salient regions. Within the CIR framework, this approach effectively distills representations derived from the full image into those predicted from the masked image, encouraging the prompted encoder to focus on salient regions rather than depending solely on global alignment. By leveraging this saliency-aware masking approach, the model's ability to generalize to novel classes and unseen data is significantly enhanced.

In addition to our proposed saliency-aware CIR framework, which introduces a novel score-based regularization, we incorporate a directional feature alignment constraint to ensure the prompted features remain aligned with the pre-trained encoder’s embeddings. To achieve this, we employ two complementary strategies. First, unlike prior works \cite{khattak2023self, zheng2023localization} that rely on sample-based alignment, we align the prompted feature with its corresponding class-wise prototype extracted from the frozen model. By using these prototypes, which represent broader class-level features, the prompted model can more effectively utilize the available data and consider diverse samples with various characteristics of the same class under the alignment constraint. Second, to maintain a balance between generalization and task adaptation, we enforce directional feature alignment rather than strict proximity. Experimental results confirm that aligning only the directional component of the prompted embeddings with the prototypes, instead of minimizing the absolute distance between embeddings that constrain both magnitude and direction, allows the model greater flexibility to adapt while preserving its generalization capabilities. This directional alignment, combined with robust prototypes, enhances the prompted model's ability to generalize to unseen classes and datasets. In summary, the main contributions of this work include:
 \begin{itemize}[align=left, leftmargin=*]
\item{We introduce CIR, a novel regularization-based prompt learning framework that promotes interaction between the modality-specific branches of prompted and frozen models, enabling cooperative learning and enhancing cross-modal alignment. Additionally, CIR employs a saliency-aware masking strategy to preserve the pre-trained model’s generalization capabilities.}

\item{We introduce a novel directional regularization approach that aligns the prompted features with class-wise prototypes, represented as mean embeddings from the frozen model. By utilizing class-wise means as reliable prototypes and emphasizing feature direction alignment over strict proximity, our proposed method improves generalization by leveraging pre-trained model capabilities while avoiding limitations inherent in distance-based metrics.}

\item{We conduct extensive evaluations on 11 popular image classification benchmarks. The results demonstrate the effectiveness of DiSa in all the base-to-novel generalization, cross-dataset transfer, domain generalization, and few-shot learning settings.}
\end{itemize} 


\section{Related Work}
\subsection{Prompt Learning for Vision-Language Models}
\label{sec:2_Related Works}

VL models \cite{jia2021scaling, radford2021learning, yao2021filip} integrate both visual and textual information to produce rich multi-modal representations. These models are generally pre-trained on vast datasets; for example, CLIP \cite{radford2021learning} and ALIGN \cite{jia2021scaling} are trained on approximately 400 million and 1 billion image-text pairs, respectively. By leveraging self-supervised learning, VL models are able to construct joint vision and language representations that greatly improve their ability to learn and generalize across different modalities. This capability enables them to excel in various tasks, particularly in few-shot \cite{chen2022plot, lu2022prompt, naeem2022i2dformer} and zero-shot visual recognition \cite{radford2021learning}. However, one of the key challenges lies in adapting these large-scale pre-trained models to specialized downstream tasks without compromising their inherent generalization ability. Prompt learning has emerged as an effective technique, leveraging learnable embeddings, known as prompt tokens, that are incorporated into model inputs \cite{lu2022prompt, derakhshani2023bayesian, zhou2022conditional}. This approach offers notable advantages, including parameter efficiency and rapid convergence, making it particularly appealing for fine-tuning foundational models like CLIP \cite{radford2021learning} across both vision and VL tasks. For instance, CoOp \cite{zhou2022learning} introduced prompt learning for CLIP by optimizing continuous prompt vectors within the language branch for few-shot image recognition. Subsequently, MaPLe \cite{khattak2023maple} extended this concept by proposing a multi-modal prompt-tuning framework that enhances transferability through the hierarchical joint learning of prompts across both vision and language branches. Although prompt learning has seen remarkable advancements, overfitting remains a significant challenge, limiting its potential for new classes and domains. This work addresses this issue by introducing a novel regularization-based prompt learning approach that preserves the generalization capabilities of the pre-trained model while enabling effective adaptation to downstream tasks.

\subsection{Regularization for Prompt Learning}
Regularization techniques play a critical role in mitigating overfitting and enhancing the ability of models to generalize effectively to unseen data. These methods can be categorized into two primary groups. The first group, known as constraint-based techniques, incorporates additional restrictions into the training process, such as weight decay~\cite{loshchilov2017decoupled} and adversarial training~\cite{augustin2020adversarial}. The second group, termed input and parameter modification strategies, includes methods like dropout~\cite{srivastava2014dropout}, model ensembling~\cite{ilharco2022patching}, label smoothing~\cite{ilharco2022patching}, and data augmentation~\cite{choi2022tokenmixup}. In the specific context of prompt learning for vision-language models, overfitting is one of the most challenging problems. Some recent works have explored self-regularization strategies that consider constraints to preserve the generalization potential of pre-trained models while fine-tuning them for downstream tasks. For instance, CoPrompt \cite{roy2023consistency} introduces a consistency constraint that aligns predictions between the prompted and pre-trained models, thus mitigating overfitting in downstream tasks. Similarly, PromptSRC \cite{khattak2023self} integrates feature-based regularization alongside the prediction-based constraint to enhance model robustness. However, these approaches often treat vision and text encoders separately, limiting cross-modal interactions and overlooking the individual contributions of each modality. To address this, we propose a novel prompt learning framework that enhances generalization by integrating cross-interactive regularization for improved cross-modal learning and directional regularization for embedding alignment with class-wise prototypes.

\definecolor{customblue}{HTML}{007FFF}
\definecolor{custompink}{HTML}{FF3399}
\begin{figure*}[t]

    \centering 
    \includegraphics[scale=0.77]{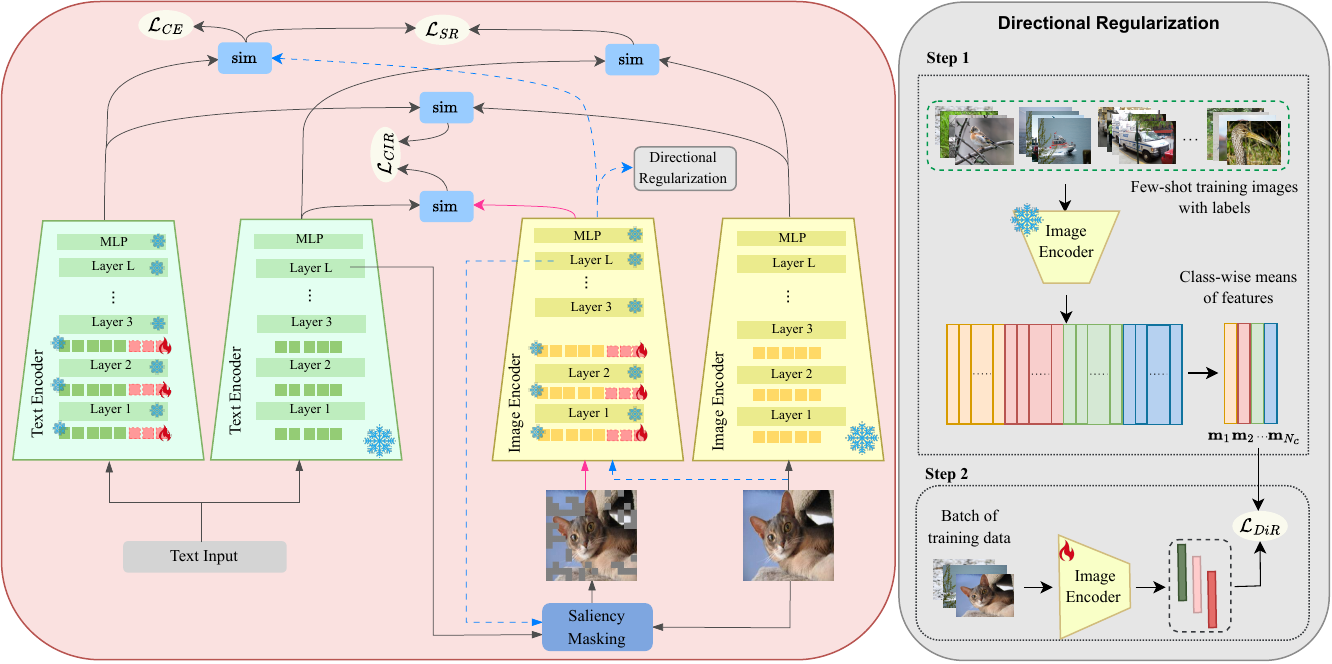}
    \caption{Overview of the proposed DiSa. The DiSa employs two complementary regularization approaches: saliency-aware cross-interactive regularization and directional regularization. The saliency-aware cross-interactive framework ensures that the prompted encoders establish independent, saliency-based interactions with the cross-modality outputs of the frozen encoders. Meanwhile, the directional regularization aligns prompted features with classification goals using class-wise feature means from the frozen model as robust prototypes. Note that the saliency masking component evaluates the importance of image tokens by computing attention scores between image patch tokens and the $CLS$ token from the frozen model's text encoder. For visual clarity, it should also be noted that we employ distinct arrow styles to illustrate different data flows: \textcolor{customblue}{dashed blue} arrows indicate the path of the full (unmasked) image through the prompted vision encoder; \textcolor{custompink}{solid pink} arrows represent the saliency-masked image path; and solid black arrows denote flows without multiple input dependencies.}
    \label{fig:Proposed Network}%
\end{figure*}


\section{Method}
\label{sec:3_Method}
\subsection{Preliminaries}\label{sec:CLIP_}
\noindent\textbf{{Contrastive Language-Image Pre-training (CLIP).}}
The CLIP model, as described in \cite{radford2021learning}, is a pre-trained system designed to align visual and textual representations by leveraging large-scale image-text datasets from the web. It is composed of two primary components: an image encoder $f(.)$ and a text encoder $g(.)$. The pre-trained model’s parameters are expressed as $\theta_{CLIP} = {\theta_{f}, \theta_{g}}$, where $\theta_{f}$ represents the parameters of the image encoder, and $\theta_{g}$ represents those of the text encoder. For any input image $x$, the feature extraction process starts by partitioning the image into $V$ fixed-size patches, which are subsequently projected into the patch feature space. A learnable class token, represented as $CLS$, is then appended to these features, forming the sequence $X = \{CLS, e_{1}, e_{2}, \dots, e_{V}\}$. This sequence is passed through $L$ layers of a transformer network, producing a visual feature representation $\mathbf{f} \in \mathbb{R}^d$. For the corresponding class label $y$, it is embedded within a text template, such as ``a photo of a {class name}'', represented as $Y = \{t_{SOS}, t_{1}, t_{2},..., t_{T}, c_{k}, t_{EOS}\}$, where $c_{k}$ is the word embedding for the class label, and $t_{SOS}$ and $t_{EOS}$ are learnable start and end tokens, respectively. Similarly to the image encoder, these text embeddings are processed through multiple transformer layers to generate the latent text feature $\mathbf{g} \in \mathbb{R}^d$. 
For zero-shot inference, text features of the text template with class labels $\{1, 2,..., N_c\}$ are matched with the image feature as $\frac{\exp(\operatorname{sim}({\mathbf{g}}^i, \mathbf{f}) / \tau)}{\sum_{i=1}^{N_c} \exp(\operatorname{sim}(\mathbf{g}^i, \mathbf{f}) / \tau)}$, where $\operatorname{sim}()$ represents cosine similarity, and $\tau$ is the temperature parameter of the softmax function.

\noindent\textbf{{Prompt Learning for CLIP.}} Inspired by prompt learning in natural language processing, numerous studies have explored the VL models by incorporating learnable prompt tokens during end-to-end training. In this study, we utilize hierarchical learnable prompt tokens independently for the text and image encoders, following the simple baseline method called Independent Vision-Language Prompting (IVLP) \cite{rasheed2023fine}. 
We concatenate learnable language prompts, denoted as $P_t = \left\{ p_t^1, p_t^2, \cdots, p_t^T \right\}$, and visual prompts, denoted as $P_v = \left\{ p_v^1, p_v^2, \cdots, p_v^V \right\}$, into the respective sets of textual and visual input tokens. Thus, the image encoder generates the prompted visual feature $\mathbf{f}_p = f(X_p, \theta_f)$ from input tokens $X_p = \{P_v, e_{CLS}, e_1, e_2, \cdots, e_{V}\}$, and the textual feature $\mathbf{g}_p = g(Y_p, \theta_g)$ is obtained from $Y_p = \{t_{SOS},\allowbreak P_t,\allowbreak t_1,\allowbreak t_2,\allowbreak \cdots,\allowbreak t_T,\allowbreak c_k,\allowbreak t_{EOS}\}$. It should be noted that in this work, we employ deep prompting, which involves learning distinct sets of prompts for each transformer layer. For image classification on a downstream dataset $D_s$, consisting of an $K$-shot $N_c$-class training set, prompts interact with frozen $\theta_f$ and $\theta_g$ and are optimized with the cross-entropy loss as:
\begin{equation}
\mathcal{L}_{CE} = \frac{1}{K{N_c}} \sum_{(x,y) \in \mathcal{D}_s} \frac{\exp \left( \operatorname{sim}({\mathbf{f}}_p, {\mathbf{g}}_{p}^{y}) / \tau \right)}{\sum_{i=1}^{N_c} \exp \left( \operatorname{sim}({\mathbf{f}}_p, {\mathbf{g}}_{p}^{i}) / \tau \right)}, 
\end{equation}
where ${\mathbf{f}}_{p}$ and ${\mathbf{g}}_{p} \in \mathbb{R}^d $ represent the vision and text embeddings derived from the prompted model, respectively.

\subsection{Directional Saliency-Aware Prompt Learning} \label{D}
As illustrated in Fig. \ref{fig:Proposed Network}, our proposed framework incorporates two complementary regularization approaches: saliency-aware cross-interactive regularization and directional regularization. The subsequent subsections provide a detailed explanation of each.
\subsubsection{Saliency-aware Cross-Interactive Regularization.}

Training prompts solely with a supervised loss specific to the task tends to undermine the general features encapsulated within the frozen CLIP model. As a result, although the weights of the CLIP image and text encoders remain unchanged, their effectiveness on unseen domains diminishes. To address this challenge, we introduce a cross-interactive regularization framework that forces prompted encoders to independently interact with the cross-modality from the pre-trained encoders. This guidance from the cross-modality helps the model to preserve its inherent generalization while adapting prompts to downstream tasks.
Additionally, benefiting from its large-scale training data, the frozen image encoder can identify and prioritize key patches for decision-making. In contrast, the prompted encoder, constrained by limited training data, risk over-relying on specific patches, resulting in significant performance degradation when encountering unseen classes or domains. To mitigate this issue, we enforce the prompted image encoder to focus on critical patches under the guidance of the frozen model. This saliency-aware regularization enhances the model's ability to generalize effectively across novel classes and diverse domains.

\noindent\textbf{{Cross-Interactive Regularization (CIR).}}
Our proposed CIR introduces an innovative method to regularize the prompted model with the frozen CLIP model by leveraging interactive learning across both visual and textual encoders. Unlike recent regularization-based methods that focus primarily on maximizing score-based similarity between the prompted and frozen models \cite{roy2023consistency}, CIR fosters a deeper interaction between the modality-specific branches of each model. This cross-interaction mechanism in CIR enables cooperative learning between the two models' modality branches, where the image embeddings from the prompted model interact with the text embeddings from the frozen model, and the text embeddings from the prompted model interact with the image embeddings of the frozen model. This structure is designed to optimize the mutual information shared across models, allowing the prompted model to capture cross-modal dependencies more effectively.
For this purpose, CIR employs  Kullback-Leibler (KL) divergence to align two critical probability scores. The first score, $q^{{{\mathbf{f}}_p} {\mathbf{g}}_o}$, is computed as the cosine similarity between the image embeddings from the prompted model and text embeddings from the frozen model. The second score, $q^{{\mathbf{f}}_o {\mathbf{g}}_p}$, is derived from the cosine similarity between the text embeddings from the prompted model and image embeddings from the frozen model. We define the CIR loss as:
\begin{equation}
\mathcal{L}_{CIR} = \mathcal{D}_{\text{KL}}(q^{{{\mathbf{f}}_p} {\mathbf{g}}_o},q^{{\mathbf{f}}_o {\mathbf{g}}_p}),
\end{equation}
\begin{equation}
q^{\mathbf{f}_p \mathbf{g}_o} =  \operatorname{sim}({\mathbf{f}}_p,{\mathbf{g}}_o), \: q^{\mathbf{f}_o \mathbf{g}_p} =  \operatorname{sim}({\mathbf{f}_o}, {\mathbf{g}_p}),  
\end{equation}
\noindent where the function $\operatorname{sim}$ represents cosine similarity. 

Furthermore, to ensure consistency between the prompted text and image encoders, we align the score predictions of the prompted model ($q^{\mathbf{f}_p \mathbf{g}_p}$) with those of the frozen CLIP model ($q^{{\mathbf{f}}_o \mathbf{g}_o}$) using the original image as follows:
\begin{equation}
\mathcal{L}_{SR} = \mathcal{D}_{\text{KL}}(q^{{\mathbf{f}}_p {\mathbf{g}}_p},q^{\mathbf{f}_o \mathbf{g}_o}),
\end{equation}
\begin{equation}
q^{\mathbf{f}_p \mathbf{g}_p} =  \operatorname{sim}({\mathbf{f}}_p, {\mathbf{g}}_p), \: q^{\mathbf{f}_o \mathbf{g}_o} =  \operatorname{sim}({\mathbf{f}_o}, {\mathbf{g}_o}). 
\end{equation}
This integrated alignment approach enables a more sophisticated cross-modal interaction, ultimately enhancing the generalizability of the prompted model across both visual and textual modalities.\\

\noindent\textbf{{Saliency Masking.}}
To ensure the prompted image encoder focuses on critical patches rather than over-relying on specific regions, we leverage the frozen model's guidance to randomly mask less important patches, providing the masked image as input to the prompted model. The $CLS$ token from the text encoder captures the semantic content of the text, and its attention weights on image tokens serve as an effective indicator of each token’s relevance to the linguistic semantics. Therefore, to identify the importance of different image tokens, we compute the attention value between the image patch tokens and the $CLS$ token from the frozen model's text encoder. By masking random, less important patches in the input image, the prompted model is encouraged to produce embeddings aligned with the frozen model, which has access to the full image. This approach encourages the model to focus on essential image regions, enhancing its ability to generalize by reducing reliance on less informative features. As a result, the model becomes more adaptable to varying data distributions and better equipped to handle challenging scenarios involving domain shifts or novel classes.

Considering the final layer of the prompted model, let \( z(n) \in \mathbb{R}^d\) denote the embedding of the \( n \)-th image patch from the image encoder, and let \( z ({CLS}) \in \mathbb{R}^d \) represent the embedding of the \( CLS \) token from the text encoder of the frozen model. The importance of each token is computed as:
\begin{equation}
\label{score}
    \alpha_n = \frac{1}{H} \sum_{h=1}^{H} \text{Softmax} \left( \frac{z_{h}^Q(\text{CLS}) \cdot z_{h}^\kappa(n)}{\sqrt{C}} \right),
\end{equation}
\noindent where $h$ indicates the attention head index; $z_{h}^Q(\text{CLS})$ denotes the query embedding of the
$CLS$ token at head $h$; $z_{h}^\kappa(n)$ is the key embedding of the $n$-th image token at head $h$, and $C$ is the dimensionality of the query and key embeddings. Image tokens to mask are selected based on the scores $\alpha_n$. We randomly mask $\gamma$ percent of the image tokens with the lowest scores.

\subsubsection{Directional Regularization.}
To complement our proposed saliency-aware CIR framework, we incorporate a directional feature alignment constraint to ensure the prompted features remain aligned with the pre-trained model’s embeddings. This is achieved through two synergistic approaches. In contrast to previous methods \cite{khattak2023self, zheng2023localization} that rely on sample-based alignment, we align the prompted feature with the corresponding prototype derived from the frozen model. These prototypes capture broader class-level features, enabling the prompted model to leverage diverse samples more effectively under the alignment constraint. To maintain generalization while enhancing adaptability, our approach aligns feature directions instead of enforcing strict proximity. Our findings indicate that aligning only the directional component of prompted embeddings with prototypes, rather than constraining both magnitude and direction, enables greater flexibility for adaptation. Thus, we employ a cosine similarity-based loss to align the direction of the prompted feature $\mathbf{f}_p$ with its corresponding class-wise prototype $\mathbf{m}_i$. Accordingly, the directional regularization loss is defined as:
\begin{equation}\label{eqoo}
\mathcal{L}_{DiR} = \left| 1 - \cos(\mathbf{f}_{p}, \mathbf{m}_i) \right|,
\end{equation}
\noindent where $\mathbf{m}_i$ represents the class-mean embedding for the $i^{\text{th}}$ class, calculated as:
\begin{equation}
\mathbf{m}_i = \frac{1}{|\textit{I}_j|} \sum_{j \in \textit{I}_j} \mathbf{f}_{o_j},
\end{equation}
\noindent with $\textit{I}_j$ denotes the set of indices corresponding to training examples from class $i$.
Finally, we employ a weighted combination of all specified loss terms for end-to-end training. Consequently, our final loss function is defined as:
\begin{equation}
\mathcal{L}_{total} = \mathcal{L}_{CE} + \mathcal{L}_{SR} +
{\mathcal{L}_{CIR}} + \lambda{\mathcal{L}_{DiR}},
\end{equation}
\noindent where $\lambda$ controls the contribution of the directional regularization loss to the total loss function.

\section{Experiments}
\subsection{Datasets and Implementation Details }
Consistent with prior research on prompt tuning \cite{rasheed2023fine, khattak2023maple, zhou2022conditional}, we assess our proposed method across four evaluation scenarios: generalization from base-to-novel classes, cross-dataset evaluation, domain generalization, and few-shot image recognition. Our experiments incorporate a variety of datasets, including two generic-object datasets (ImageNet \cite{deng2009imagenet} and Caltech101 \cite{fei2004learning}), fine-grained datasets (OxfordPets \cite{parkhi2012cats}, StanfordCars \cite{krause20133d}, Flowers102 \cite{nilsback2008automated}, Food101 \cite{bossard2014food}, and FGVCAircraft\cite{maji2013fine}), a remote sensing classification dataset (EuroSAT\cite{helber2019eurosat}), a scene recognition dataset (SUN397 \cite{xiao2010sun}), an action recognition dataset (UCF101\cite{soomro2012ucf101}), and a texture dataset (DTD\cite{cimpoi2014describing}). For domain generalization, we employ ImageNet as the source dataset, and its four variants as target datasets, including ImageNetV2 \cite{recht2019imagenet}, ImageNetSketch \cite{wang2019learning}, ImageNet-A\cite{hendrycks2021natural}, and ImageNet-R \cite{hendrycks2021many}.

\begin{table*}[t]
    \centering
\footnotesize
\caption{Comparison of different methods in 16-shot base-to-novel generalization. We report the accuracy (\%) on both base and novel classes, as well as their harmonic mean. The best results are presented in bold.}
    \begin{subfigure}[t]{0.32\textwidth}
        \centering
        \caption*{(a) Average over 11 datasets}
        \begin{tabular}{lccc}
            \toprule
            & Base & New & HM \\
            \midrule
            CLIP  \cite{radford2021learning} & 69.34 & 74.22 & 71.70 \\
            CoOp \cite{zhou2022learning} & 82.69 & 63.22 & 71.66 \\
            CoCoOp \cite{zhou2022conditional} & 80.47 & 71.69 & 75.83 \\
            Maple \cite{khattak2023maple} & 82.28 & 75.14 & 78.55 \\
              PromptSRC \cite{khattak2023self} & 84.26 & 76.10 & 79.97 \\
                CoPrompt \cite{roy2023consistency} & 84.00 & 77.23 & 80.48 \\
                  MMA \cite{yang2024mma} & 83.20 & 76.80 & 79.87 \\
                   APEX \cite{yang2023towards} & 83.99 & 76.76 & 80.04 \\
                     TCP \cite{yao2024tcp}& 84.13 & 75.36 & 79.51 \\
             \hdashline [2pt/1.5pt]
            DiSa (Ours) & \textbf{84.43} & \textbf{78.05} & \textbf{81.11} \\
            \bottomrule
        \end{tabular}
    \end{subfigure}%
    \hfill
    \vspace{-0.1cm}
    \begin{subfigure}[t]{0.32\textwidth}
        \centering
        \caption*{(b) ImageNet}
        \begin{tabular}{lccc}
            \toprule
            & Base & New & HM \\
            \midrule
            CLIP\cite{radford2021learning} & 72.43 & 68.14 & 70.22 \\
            CoOp\cite{zhou2022learning} & 76.47 & 67.88 & 71.92 \\
            CoCoOp\cite{zhou2022conditional} & 75.98 & 70.43 & 73.10 \\
            Maple \cite{khattak2023maple} &76.66 & 70.54 & 73.47 \\
              PromptSRC \cite{khattak2023self} & {77.60} & 70.73 & 74.01 \\
                CoPrompt \cite{roy2023consistency} & \textbf{77.67} & 71.27 & 74.33 \\
                  MMA \cite{yang2024mma} & 77.31 & 71.00 & 74.02 \\
                   APEX \cite{yang2023towards} & 77.12 & 71.10 & 73.99 \\
                     TCP \cite{yao2024tcp}& 77.27 & 69.87 & 73.38 \\
             \hdashline [2pt/1.5pt]
            DiSa (Ours) & {77.56} & \textbf{71.65} & \textbf{74.49} \\
            \bottomrule
        \end{tabular}
    \end{subfigure}%
    \hfill
     \vspace{-0.1cm}
    \begin{subfigure}[t]{0.32\textwidth}
        \centering
        \caption*{(c) Caltech101}
        \begin{tabular}{lccc}
            \toprule
            & Base & New & HM \\
            \midrule
            CLIP\cite{radford2021learning} & 96.84 & 94.00 & 95.40 \\
            CoOp\cite{zhou2022learning} & 98.00 & 89.81 & 93.73 \\
            CoCoOp \cite{zhou2022conditional}& 97.96 & 93.81 & 95.84 \\
            Maple \cite{khattak2023maple} & 97.74 & 94.36 & 96.02 \\
              PromptSRC \cite{khattak2023self} & 98.10 & 94.03 & 96.02 \\
                CoPrompt \cite{roy2023consistency} & 98.27 & 94.90 & 96.55 \\
                  MMA \cite{yang2024mma} & \textbf{98.40} & 94.00 & 96.15 \\
                   APEX \cite{yang2023towards} & 98.18 & 95.06 & 96.59 \\
                     TCP \cite{yao2024tcp}& 98.23 & 94.67 & 96.42 \\
             \hdashline [2pt/1.5pt]
            DiSa (Ours) & {98.29} & \textbf{95.41} & \textbf{96.83} \\
            \bottomrule
        \end{tabular}
    \end{subfigure}%
    
    \vspace{0.3cm}
    
    \begin{subfigure}[t]{0.32\textwidth}
        \centering
        \caption*{(d) OxfordPets}
        \begin{tabular}{lccc}
            \toprule
            & Base & New & HM \\
            \midrule
            CLIP\cite{radford2021learning} & 91.17 & 97.26 & 94.12 \\
            CoOp\cite{zhou2022learning} & 93.67 & 95.29 & 94.47 \\
            CoCoOp\cite{zhou2022conditional} & 95.20 & 97.69 & 96.43 \\
            Maple \cite{khattak2023maple} & 95.43 & 97.76 & 96.58 \\
              PromptSRC \cite{khattak2023self} &95.33 & 97.30 & 96.30 \\
                CoPrompt \cite{roy2023consistency} & \textbf{95.67} & 98.10 & 96.87 \\
                  MMA \cite{yang2024mma} & 95.40 & 98.07 & 96.72 \\
                   APEX \cite{yang2023towards} & 95.11 & 97.27 & 96.18 \\
                     TCP \cite{yao2024tcp}& 94.67 & 97.20 & 95.92 \\
             \hdashline [2pt/1.5pt]
            DiSa (Ours) & {95.48} & \textbf{98.67} & \textbf{97.05} \\
            \bottomrule
        \end{tabular}
    \end{subfigure}%
    \hfill
     \vspace{-0.1cm}
    \begin{subfigure}[t]{0.32\textwidth}
        \centering
        \caption*{(e) StanfordCars}
        \begin{tabular}{lccc}
            \toprule
            & Base & New & HM \\
            \midrule
            CLIP\cite{radford2021learning} & 63.37 & 74.89 & 68.65 \\
            CoOp\cite{zhou2022learning} & 78.12 & 60.40 & 68.13 \\
            CoCoOp\cite{zhou2022conditional} & 70.49 & 73.59 & 72.01 \\
                        Maple \cite{khattak2023maple} & 72.94 & 74.00 & 73.47 \\
              PromptSRC \cite{khattak2023self} & 78.27 & 74.97 & 76.58 \\
                CoPrompt \cite{roy2023consistency} & 76.97 & 74.40 & 75.66 \\
                  MMA \cite{yang2024mma} & {78.50} & 73.10 & 75.70 \\
                   APEX \cite{yang2023towards} & 80.53 & \textbf{75.08} & \textbf{77.71} \\
                     TCP \cite{yao2024tcp}& \textbf{80.80} & 74.13 & 77.32 \\
             \hdashline [2pt/1.5pt]
            DiSa (Ours) & {78.54} & {75.07} & {76.77} \\
            \bottomrule
        \end{tabular}
    \end{subfigure}%
    \hfill
     \vspace{-0.1cm}
    \begin{subfigure}[t]{0.32\textwidth}
        \centering
        \caption*{(f) Flowers102}
        \begin{tabular}{lccc}
            \toprule
            & Base & New & HM \\
            \midrule
            CLIP\cite{radford2021learning} & 72.08 & 77.80 & 74.83 \\
            CoOp\cite{zhou2022learning} & 97.60 & 59.67 & 74.06 \\
            CoCoOp\cite{zhou2022conditional} & 94.87 & 71.75 & 81.71 \\
            Maple \cite{khattak2023maple} & 95.92 & 72.46 & 82.56 \\
              PromptSRC \cite{khattak2023self} & {98.07} & 76.50 & 85.95 \\
                CoPrompt \cite{roy2023consistency} & 97.27 & 76.60 & 85.71 \\
                  MMA \cite{yang2024mma} & 97.77 & 75.93 & 85.48 \\
                   APEX \cite{yang2023towards} & 97.47 & \textbf{77.58} & \textbf{86.40} \\
                     TCP \cite{yao2024tcp}& 97.73 & 75.57 & 85.23 \\
             \hdashline [2pt/1.5pt]
            DiSa (Ours) & \textbf{98.14} & {76.77} & {86.15} \\
            \bottomrule
        \end{tabular}
    \end{subfigure}%

    \vspace{0.3cm}
    \begin{subfigure}[t]{0.32\textwidth}
        \centering
        \caption*{(g) Food101}
        \begin{tabular}{lccc}
           \toprule
            & Base & New & HM \\
            \midrule
            CLIP\cite{radford2021learning} & 92.43 & 91.22 & 90.66 \\
            CoOp\cite{zhou2022learning} & 88.33 & 82.26 & 85.19 \\
            CoCoOp\cite{zhou2022conditional} & 90.70 & 91.29 & 90.99 \\
            Maple \cite{khattak2023maple} & 90.71 & 92.05 & 91.38 \\
              PromptSRC \cite{khattak2023self} & 90.67 & 91.53 & 91.10 \\
                CoPrompt \cite{roy2023consistency} & 90.73 & 92.07 & 91.40 \\
                  MMA \cite{yang2024mma} & 90.13 & 91.30 & 90.71 \\
                   APEX \cite{yang2023towards} & 89.60 & 92.06 & 90.81 \\
                     TCP \cite{yao2024tcp}& 90.57 &91.37 & 90.97 \\
             \hdashline [2pt/1.5pt]
            DiSa (Ours) & \textbf{90.81} & \textbf{92.32} & \textbf{91.56} \\
            \hline 
        \end{tabular}
    \end{subfigure}%
    \hfill
     \vspace{-0.1cm}
    \begin{subfigure}[t]{0.32\textwidth}
        \centering
        \caption*{(h) FGVCAircraft}
        \begin{tabular}{lccc}
            \toprule
            & Base & New & HM \\
            \midrule
            CLIP\cite{radford2021learning} & 27.19 & 36.29 & 31.09 \\
            CoOp\cite{zhou2022learning} & 40.44 & 22.30 & 28.75 \\
            CoCoOp\cite{zhou2022conditional} & 33.41 & 23.71 & 27.74 \\
            Maple \cite{khattak2023maple} & 37.44 & 35.61 & 36.50 \\
              PromptSRC \cite{khattak2023self} & \textbf{42.73} & 37.87 & 40.15 \\
                CoPrompt \cite{roy2023consistency} & 40.20 & 39.33 & 39.76\\
                  MMA \cite{yang2024mma} & 40.57 & 36.33 & 38.33 \\
                   APEX \cite{yang2023towards} & 42.69 & 35.21 & 38.59 \\
                     TCP \cite{yao2024tcp}& 41.97 & 34.43 & 37.83 \\
             \hdashline [2pt/1.5pt]
            DiSa (Ours) & {42.65} & \textbf{39.38} & \textbf{40.95} \\
            \bottomrule
        \end{tabular}
    \end{subfigure}%
    \hfill
     \vspace{-0.15cm}
    \begin{subfigure}[t]{0.32\textwidth}
        \centering
        \caption*{(i) SUN397}
        \begin{tabular}{lccc}
            \toprule
            & Base & New & HM \\
            \midrule
            CLIP\cite{radford2021learning} & 69.36 & 75.35 & 72.23 \\
            CoOp \cite{zhou2022learning}& 80.60 & 65.89 & 72.51 \\
            CoCoOp\cite{zhou2022conditional} & 79.74 & 76.86 & 78.27 \\
            Maple \cite{khattak2023maple} & 80.82 & 78.70 & 79.75 \\
              PromptSRC \cite{khattak2023self} & 82.67 & 78.47 & 80.52 \\
                CoPrompt \cite{roy2023consistency} & 82.63 & 80.03 & 81.31 \\
                  MMA \cite{yang2024mma} & 82.27 & 78.57 & 80.38 \\
                   APEX \cite{yang2023towards} & 81.17 & 78.98 & 80.06 \\
                     TCP \cite{yao2024tcp}& 82.63 & 78.20 &80.35 \\
             \hdashline [2pt/1.5pt]
            DiSa (Ours) & \textbf{82.69} & \textbf{80.53} & \textbf{81.60} \\
            \bottomrule
        \end{tabular}
    \end{subfigure}%

    \vspace{0.3cm}
    
    \begin{subfigure}[t]{0.32\textwidth}
        \centering
        \caption*{(j) DTD}
        \begin{tabular}{lccc}
            \toprule
            & Base & New & HM \\
            \midrule
            CLIP\cite{radford2021learning} & 53.24 & 59.90 & 56.37 \\
            CoOp\cite{zhou2022learning} & 79.44 & 41.18 & 54.24 \\
            CoCoOp\cite{zhou2022conditional} & 77.01 & 56.00 & 64.85 \\
                       Maple \cite{khattak2023maple} & {80.36} & 59.18 & 68.16 \\
              PromptSRC \cite{khattak2023self} & \textbf{83.37} & 62.97 & 71.75 \\
                CoPrompt \cite{roy2023consistency} & 83.13 & 64.73 & 72.79 \\
                  MMA \cite{yang2024mma} & 83.20 & 65.63 & 73.38 \\
                 
                  APEX \cite{yang2023towards}& 82.45 & 63.80 & 71.94 \\
                  TCP \cite{yao2024tcp} & 82.77 & 58.07 & 68.25 \\
             \hdashline [2pt/1.5pt]
            DiSa (Ours) & {83.33} & \textbf{65.71} & \textbf{73.49} \\
            \bottomrule
        \end{tabular}
    \end{subfigure}%
    \hfill
     \vspace{-0.1cm}
    \begin{subfigure}[t]{0.32\textwidth}
        \centering
        \caption*{(k) EuroSAT}
        \begin{tabular}{lccc}
            \toprule
            & Base & New & HM \\
            \midrule
            CLIP\cite{radford2021learning} & 56.48 & 64.05 & 60.03 \\
            CoOp\cite{zhou2022learning} & 92.19 & 54.74 & 68.69 \\
            CoCoOp \cite{zhou2022conditional}& 87.49 & 60.04 & 71.21 \\
             Maple \cite{khattak2023maple} & 94.07 & 73.23 & 82.35 \\
              PromptSRC \cite{khattak2023self} & 92.90 & 73.90 & 82.32 \\
                CoPrompt \cite{roy2023consistency} & \textbf{94.60}& 78.57 & 85.84 \\
                  MMA \cite{yang2024mma} & 85.46 & 82.34 & 83.87 \\
                   APEX \cite{yang2023towards} & 92.83 & 79.89 & 85.85 \\
                    TCP \cite{yao2024tcp}& 91.63 & 74.73 & 82.32 \\
             \hdashline [2pt/1.5pt]
            DiSa (Ours) & {94.10} & \textbf{82.69} & \textbf{88.03} \\
            \bottomrule
        \end{tabular}
    \end{subfigure}%
    \hfill
     \vspace{-0.1cm}
    \begin{subfigure}[t]{0.32\textwidth}
        \centering
        \caption*{(l) UCF101}
        \begin{tabular}{lccc}
            \toprule
            & Base & New & HM \\
            \midrule
            CLIP \cite{radford2021learning}& 70.53 & 77.50 & 73.85 \\
            CoOp \cite{zhou2022learning}& 84.69 & 56.05 & 67.46 \\
            CoCoOp\cite{zhou2022conditional} & 82.33 & 73.45 & 77.64 \\
             Maple \cite{khattak2023maple} & 83.00 & 78.66 & 80.77 \\
              PromptSRC \cite{khattak2023self} & 87.10 & 78.80 & 82.74 \\
                CoPrompt \cite{roy2023consistency} & 86.90 & 79.57 & 83.07 \\
                  MMA \cite{yang2024mma} & 86.23 & 80.03 & 82.20 \\
                   APEX \cite{yang2023towards} & 86.74 & 78.37 & 82.34 \\
                     TCP \cite{yao2024tcp}& 87.13 & \textbf{80.77} & \textbf{83.83} \\
             \hdashline [2pt/1.5pt]
            DiSa (Ours) & \textbf{87.16} & {80.45} & {83.67} \\
            \bottomrule
        \end{tabular}
    \end{subfigure}%
\vspace{0.5cm}
    
     \label{tab:base_novel}
\end{table*}

\begin{table*}[h!]
    \centering
    \footnotesize
        \caption{Comparison of our proposed method with state-of-the-art approaches in cross-dataset evaluation. Our method achieves superior average performance across 10 datasets, highlighting its strong zero-shot adaptability.}
    
   \setlength\tabcolsep{5pt}
      \resizebox{0.88\textwidth}{!}{%
    \begin{tabular}{lcccccccccccc}
    \hline \hline
    \addlinespace[0.5mm]
    & \multicolumn{1}{c}{\textbf{Source}} & \multicolumn{11}{c}{\textbf{Target}} \\
    \cmidrule(lr){2-2} \cmidrule(lr){3-13}
    & \rotatebox{45}{\textbf{ImageNet}} & \rotatebox{45}{\textbf{Caltech101}} & \rotatebox{45}{\textbf{OxfordPets}} & \rotatebox{45}{\textbf{StanfordCars}} & \rotatebox{45}{\textbf{Flowers102}} & \rotatebox{45}{\textbf{Food101}} & \rotatebox{45}{\textbf{Aircraft}} & \rotatebox{45}{\textbf{SUN397}} & \rotatebox{45}{\textbf{DTD}} & \rotatebox{45}{\textbf{EuroSAT}} & \rotatebox{45}{\textbf{UCF101}} & \rotatebox{45}{\textbf{Average}} \\
    \midrule
    \textbf{CoOp} \cite{zhou2022learning}& 71.51 & 93.70 & 89.14 & 64.51 & 68.71 & 85.30 & 18.47 & 64.15 & 41.92 & 46.39 & 66.55 & 63.88 \\
    \textbf{CoCoOp} \cite{zhou2022conditional}& 71.02 & 94.43 & 90.14 & 65.32 & 71.88 & 86.06 & 22.94 & 67.36 & 45.73 & 45.37 & 68.21 & 65.74 \\
    
    \textbf{MaPLe} \cite{khattak2023maple}& 70.72 & 93.53 & 90.49 & 65.57& 72.23& 86.20 & 24.74 & 67.01 & 46.49 & 48.06 & 68.69 & 66.30 \\

      \textbf{PromtSCR} \cite{khattak2023self}& {71.27} &93.60 &90.25& 65.70& 70.25 &86.15& 23.90 &67.10 &46.87& 45.50 &68.75 &65.81 \\

   \textbf{CoPrompt} \cite{roy2023consistency}& 70.80& 94.50 &90.73& 65.67 &72.30 &86.43& 24.00& 67.57&47.07& \textbf{51.90}& 69.73 &67.00\\
      \textbf{MMA} \cite{yang2024mma}&71.00 &93.80 &90.30& {66.13} &72.07 &86.12& \textbf{25.33} &68.17& 46.57& 49.24& 68.32& 66.61\\ 
        
  \textbf{APEX} \cite{yang2023towards}& \textbf{72.00} & 94.46 & 90.06 & 65.46 & 71.58 & 86.44 & 24.44 & 67.20 & 45.70 & 47.58 & 68.80 & 66.16 \\
   \textbf{TCP} \cite{yao2024tcp}& 71.40 & 93.97 & \textbf{91.25} & 64.69 & 71.21 & \textbf{86.69} & 23.45 & 67.15 & 44.35 & 51.45 & 68.73 & 66.29 \\
      \hdashline [2pt/1.5pt]

      \addlinespace[0.5mm]
   \textbf{DiSa}& 71.21 & \textbf{94.62}& {90.94} & \textbf{66.22} & \textbf{72.51} & {86.64} & 25.26 & \textbf{68.32} & \textbf{47.23} &50.84 & \textbf{69.93} & \textbf{67.25} \\
      \hline \hline
  
    \end{tabular}}

\label{tab:cross_comparison}
\end{table*}

In all our experiments, to align the comparisons with previous approaches \cite{khattak2023maple, khattak2023self}, we utilize a ViT-B/16-based CLIP model and employ deep prompting with \(V = T = 4\) VL prompts. It should be noted that the prompts are randomly initialized using a normal distribution, except for the text prompts in the first layer, which are initialized with the word embeddings of ``a photo of a". For domain generalization and cross-dataset evaluation, learnable prompts are injected into the first three transformer layers. For the few-shot and base-to-novel settings, prompts are injected into the first nine transformer layers. Each model is trained with a batch size of 4 and a learning rate of 0.0025 using the SGD optimizer on a single NVIDIA RTX 4090 GPU. Training is conducted for 50 epochs in the few-shot setting, while for all other settings, the model is trained for 20 epochs under a 16-shot configuration, where only 16 training samples per category are provided.
The results are averaged over three runs. The parameter $\lambda$ is set to 12. For masking less informative patches, we identify the 50\% of patches ($\gamma=$ 50\%) with the lowest scores (calculated as per Eq. \ref{score}) and randomly mask half of these less important patches.

\begin{table}[!]
    \centering
    \small
        \caption{Comparison of our proposed method with state-of-the-art studies in
the domain generalization setting.}
       \setlength\tabcolsep{3.1pt}
\resizebox{0.4\textwidth}{!}{%
    \begin{tabular}{lccccccc}
          \hline \hline
    \addlinespace[0.5mm]
        & \multicolumn{1}{c}{\textbf{Source}} & \multicolumn{5}{c}{\textbf{Target}} & \\
        \cmidrule(lr){2-2} \cmidrule(lr){3-7}
        & \textbf{ImageNet} & \textbf{-V2} & \textbf{-S} & \textbf{-A} & \textbf{-R} & \textbf{Avg.} \\
        \midrule
        \textbf{CLIP} \cite{radford2021learning}& 66.73 & 60.83 & 46.15 & 47.77 & 73.96 & 57.18 \\
        \textbf{CoOp} \cite{zhou2022learning}& 71.51 & 64.20 & 47.99 & 49.71 & 75.21 & 59.28 \\
        \textbf{CoCoOp} \cite{zhou2022conditional}& 71.02 & 64.07 & 48.75 & 50.63 & 76.18 & 59.91 \\
        \textbf{MaPLe}  \cite{khattak2023maple}& 70.72 & 64.07 & 49.15 & 50.90 & 76.98 & 60.27 \\
        \textbf{PromptSRC} \cite{khattak2023self}&{71.27} & 64.35 & 49.55 & 50.90 & 77.80 & 60.65 \\
        \textbf {CoPrompt} \cite{roy2023consistency}& 70.80 & 64.81 & 49.54 & 51.51 & 77.34 & 60.80 \\
          \textbf{MMA} \cite{yang2024mma}& 71.00 & 64.33 & 49.13 & 51.12 & 77.32 & 60.77 \\
            \textbf{APEX} \cite{yang2023towards}&  \textbf{72.00} & 64.70 & 48.48 & 50.68 & 76.76 & 60.16 \\
               \textbf{TCP} \cite{yao2024tcp}& 71.20 & 64.60 & 49.50 & 51.20 & 76.73 & 60.51 \\
        \hdashline [2pt/1.5pt]
         \addlinespace[0.5mm]
       
          \textbf{DiSa}& {71.21} & \textbf{65.60} & \textbf{50.29} & \textbf{51.90} & \textbf{77.94} & \textbf{61.43} \\
          \hline \hline
    \end{tabular}}
    \label{tab:domain}
\end{table}



\subsection{Base-to-Novel Generalization}
We evaluate the generalization ability of our proposed approach across 11 different image classification datasets. Following the experimental setup outlined in prior studies \cite{zhou2022conditional, khattak2023maple, lu2022prompt}, each dataset is divided into base and novel classes. The model is trained on the base classes using a few-shot learning strategy with 16 shots and is later tested on both the base and novel classes.
Performance results, detailed in Table \ref{tab:base_novel}, compare our proposed method with state-of-the-art methods, including zero-shot CLIP \cite{radford2021learning}, CoOp \cite{zhou2022learning}, CoCoOp \cite{zhou2022conditional}, MaPLe \cite{khattak2023maple}, CoPrompt \cite{roy2023consistency}, PromptSRC \cite{khattak2023self}, and MMA \cite{yang2024mma}, across the 11 datasets. The proposed DiSa framework delivers remarkable advancements over existing methods, with an average improvement of 0.82\% in the demanding novel-class setting. Beyond its core goal of advancing generalization to novel scenarios, DiSa demonstrates a 0.17\% average performance gain over PromptSRC for base classes, illustrating its superior adaptability while maintaining strong generalization capabilities.

\subsection{Cross-Dataset Evaluation}
We further consider a more challenging setting to assess the generalization capability of our method across different datasets. Similar to the state-of-the-art methods \cite{zhou2022conditional, lu2022prompt, roy2023consistency}, we train the model on ImageNet \cite{deng2009imagenet} and directly evaluate it on other datasets with all learnable parameters frozen. Table \ref{tab:cross_comparison} reports the results of DiSa and other methods on out-of-distribution datasets. DiSa achieves the highest average performance among all methods, with an average improvement of 0.25\%, showcasing the effectiveness of our CIR framework in enabling adaptability across diverse datasets.
\subsection{Domain Generalization}

In this evaluation setting, consistent with earlier studies \cite{zhou2022conditional, lu2022prompt, roy2023consistency, yang2024mma, roy2023consistency}, we assess the transferability of the ImageNet-trained model to several ImageNet variant datasets. As illustrated in Table \ref{tab:domain}, our proposed method outperforms existing approaches, achieving a top average accuracy of 61.43\%. These findings highlight the impact of cross-interactive regularization and saliency guidance in enabling DiSa to generalize effectively across out-of-distribution domains.
\vspace{-2pt}

\begin{figure*}[t] 

    \centering 
    \includegraphics[scale=0.50]{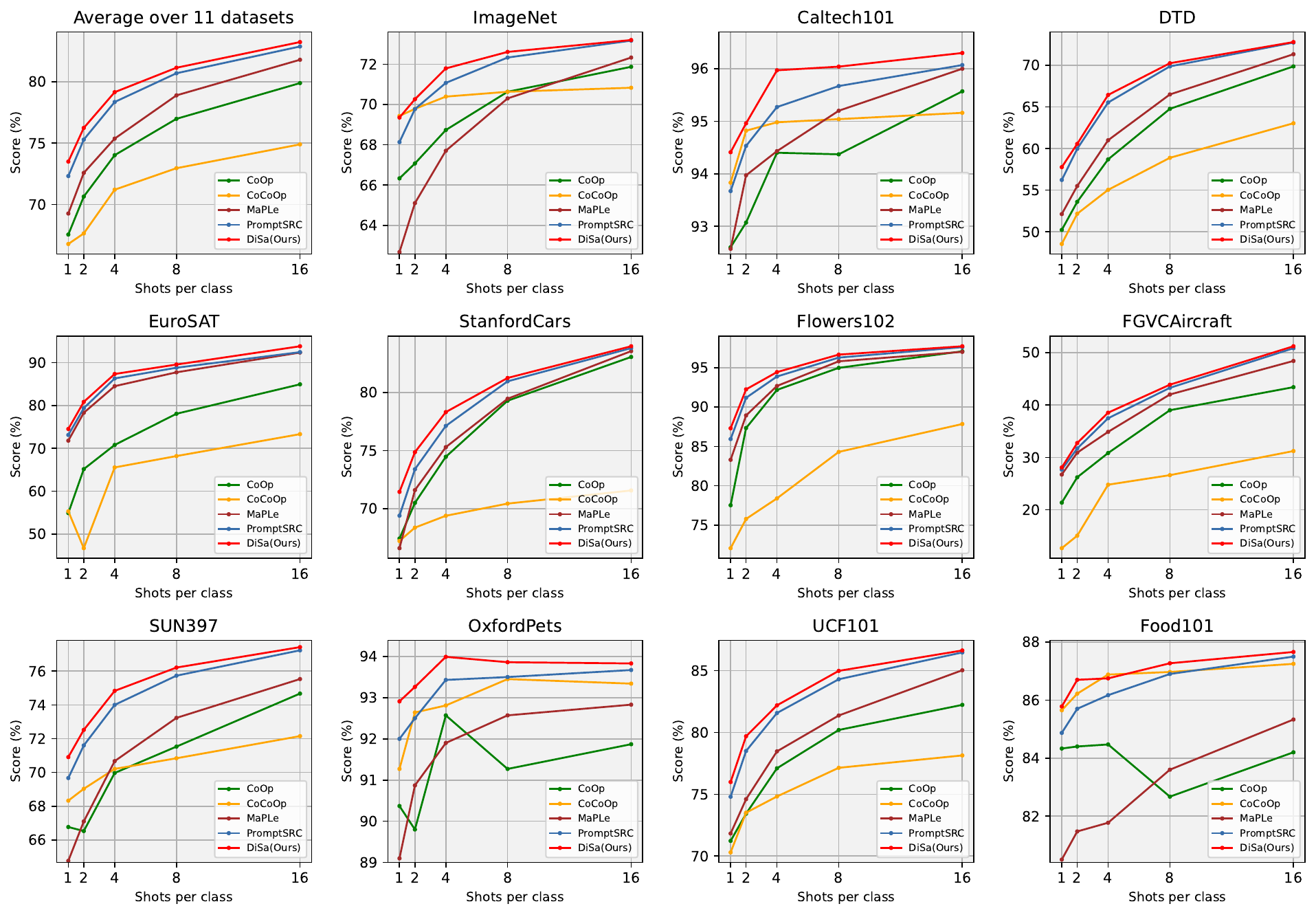}
    \caption{Performance comparison across K-shot settings (K = 1, 2, 4, 8, 16). Our approach consistently achieves superior average performance, with notable gains in low-shot scenarios, particularly for K=1, 2, 4. }
    \label{fig:fewshot}%
\end{figure*}

\subsection{Few-shot Experiments}
We further extend our experiments to the few-shot learning benchmark to assess the impact of our regularization framework on the ability of prompts to acquire task-specific knowledge. Figure \ref{fig:fewshot} presents a comparative analysis of our proposed method against existing approaches across various K-shot settings (K = 1, 2, 4, 8, 16). As observed, our method consistently outperforms existing methods in terms of average performance, achieving gains of 1.18\%, 0.95\%, 0.80\%, 0.45\%, and 0.36\% for 1, 2, 4, 8, and 16 shots across 11 datasets. Furthermore, our approach exhibits relatively larger performance gains in low-shot scenarios, particularly for K=1, 2, and 4 across almost all datasets. This indicates that DiSa effectively mitigates overfitting while allowing the prompts to capture task-specific knowledge.

\subsection{Ablation and Analysis}

\noindent\textbf{Effectiveness of regularization constraints.}
To clarify the impact of our various regularization constraints, we perform a series of experiments. The first row in Table \ref{tab:ablation} presents the performance of the baseline, which corresponds to the IVLP prompt learning approach. In the second through fourth rows, we apply the cross-interactive regularization, cross-interactive regularization with masking strategies, and score-based regularizations, respectively. The final two rows assess the contribution of the proposed directional regularization. Specifically, in the penultimate row, we align the feature direction of the prompted model with that of the frozen model derived from a single sample. In contrast, the final row aligns the prompted features with class-wise prototypes. Comparing these two settings highlights the considerable contribution of employing class-wise prototypes, which effectively adapt the model for both base and novel classes.  The results clearly show that each regularization constraint plays a vital role in enhancing the model’s generalization, leading to substantial improvements for novel classes. Furthermore, while the primary goal is to boost generalization capabilities, the integration of CIR and robust prototypes with directional alignment also delivers a measurable 0.22\% improvement for base classes over the baseline, further demonstrating the overall robustness of the proposed approach.

\definecolor{lightgray}{gray}{0.3}

\begin{table}[t!]
    \centering
    \caption{Analysis of the effectiveness of each component in DiSa. $\bm{\mathcal{L}_{DiR}}$ with \textcolor{lightgray} {\textit{Sample}} is included for comparison only.}
    \footnotesize
    \setlength\tabcolsep{2.pt}
    \renewcommand{\arraystretch}{1.2} 
\resizebox{0.36\textwidth}{!}{%
    \begin{tabular}{l c c c| >{\color{lightgray}} c  c| c c c}
        \hline \hline
        & \multicolumn{5}{c|}{\textbf{Approach}} & \multicolumn{3}{c}{\textbf{Accuracy}} \\
        \hline
        & \multirow{2}{*}{$\bm{\mathcal{L}_{CIR}}$} & \multirow{2}{*}{$\bm{Masking}$} & \multirow{2}{*}{$\bm{\mathcal{L}_{SR}}$} & \multicolumn{2}{c|}{$\bm{\mathcal{L}_{DiR}}$} 
        & \multirow{2}{*}{\textbf{Base}} & \multirow{2}{*}{\textbf{Novel}} & \multirow{2}{*}{\textbf{HM}} \\
        & & & & \textit{Sample} & \textit{Prototype}& & & \\ 
        \hline
        &  &  &  &  &  & 84.21 & 71.79 & 77.51 \\
        \arrayrulecolor{lightgray} \hline
        & $\checkmark$ &  &  &  &  & 84.35 & 74.23 & 78.97 \\
        \hline
        & $\checkmark$  & $\checkmark$  &  &  &  & 84.31 & 74.86 & 79.30 \\
        \hline
        & $\checkmark$  & $\checkmark$  & $\checkmark$ &  &  & 84.27 & 76.53 & 80.21 \\
        \hline
                & $\checkmark$  & $\checkmark$  & $\checkmark$ & $\checkmark$  &  & 84.25 & 77.09 & 80.51 \\
                \hline
        & $\checkmark$  & $\checkmark$  & $\checkmark$ &  & $\checkmark$ & \textbf{84.43} & \textbf{78.05} & \textbf{81.11} \\
        \arrayrulecolor{black} \hline \hline
    \end{tabular}}
    \label{tab:ablation}
\end{table}

\begin{figure*}[t] 

    \centering 
    \includegraphics[scale=0.38]{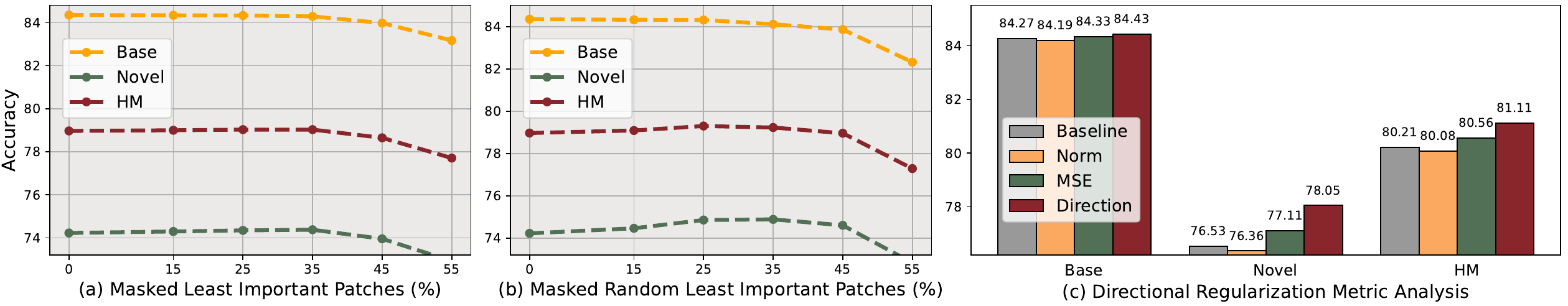}
    \caption{Analysis of saliency masking and the directional regularization. (a) Accuracy vs. percentage of least informative patches masked, showing optimal performance for novel classes at 25–35\% masking. (b) Random masking of 50\% least informative patches improves novel-class accuracy, with stability for base classes at 25\% masking. (c) Comparison of feature alignment strategies, highlighting directional alignment as most effective for improving generalization.}
    \label{fig:ablation}%
\end{figure*}

\noindent\textbf{Analysis of the saliency-masking approach.} The results in Table \ref{tab:ablation} demonstrate that integrating the proposed saliency-masking approach with the CIR framework significantly enhances model performance. This integration achieves over a 0.63\% improvement for novel classes, which pose the greatest challenge. In this section, we provide a detailed analysis of the proposed saliency-masking approach and explore the impact of its hyper-parameters. To this end, we conduct two sets of experiments to evaluate the impact of masking on model performance. In the first set (Fig. \ref{fig:ablation} (a)), we investigate the effect of masking different percentages of the input image by selectively masking the least informative patches. The results indicate a slight improvement for novel classes when 25\% to 35\% of the patches are masked. However, masking less than 25\% yields negligible changes in performance, while masking more than 35\% leads to performance degradation for both base and novel classes. Therefore, moderate masking levels (25–35\%) effectively guide the prompted image encoder to attend to the most relevant regions. However, increasing the masking ratio beyond this range starts to eliminate patches containing important contextual cues, which in turn disrupts the model’s ability to capture meaningful patterns. In the second set of experiments (Fig. \ref{fig:ablation} (b)), we extend the masking strategy by first identifying the least informative patches (defined as the bottom 50\% in importance, as described in Subsection \ref{D}) and then randomly masking a specific percentage of these patches. The results indicate that masking 25\% of the least informative patches (equivalent to 50\% masking within this subset) significantly enhances performance for novel classes while maintaining stable performance for base classes. These findings validate the effectiveness of randomness in patch selection.

\noindent\textbf{Analysis of directional regularization.}
To clarify the motivation for aligning the prompted feature direction with class-mean features, we conduct additional experiments. Each feature vector, $\mathbf{f} \in \mathbb{R}^d$, can be considered as consisting of two main components: the norm ($||\mathbf{f}||_2$) and the direction ($\mathbf{f}/||\mathbf{f}||_2$) components. Based on this, we define three types of alignment. The first involves aligning the norm components by minimizing the distance $|\|\mathbf{f}\|_2 - \|\mathbf{m}_i\|_2|$. The second addresses the alignment of the entire feature by minimizing the MSE loss $(1/d) \|\mathbf{f} - \mathbf{m}_i\|_2^2$ which includes both norm and direction information. The third focuses exclusively on aligning the directional components, as formalized in Eq. \ref{eqoo}. As shown in Fig. \ref{fig:ablation} (c), norm alignment degrades performance on novel classes, while directional alignment yields the most improvement. This underscores the importance of directional alignment in preventing overfitting and enhancing generalization.

\noindent\textbf{Sensitivity study.} This section presents a sensitivity analysis of the proposed method with respect to key hyperparameters. We first examine the influence of the weighting factor, $\lambda$, on model performance. Given that the magnitude of the $\mathcal{L}_{\text{DiR}}$ loss is considerably smaller than that of other components in the total loss function, we assign it a higher weight to ensure its effective contribution to the overall optimization objective. Table \ref{lambda} summarizes the impact of varying $\lambda$ values on the average performance across 11 datasets. The results clearly indicate that increasing $\lambda$ improves performance, with the best average accuracy across all datasets achieved at $\lambda = 12$.

Additionally, we analyze the impact of the number of prompted layers on DiSa's performance. As summarized in Table \ref{pdepth}, injecting learnable prompts into the first three transformer layers yields the best results for domain generalization setting. In the case of base-to-novel setting, the best performance is achieved by injecting learnable prompts into the first nine layers. These findings demonstrate that our model maintains stable and robust performance across a range of hyperparameter values, highlighting its practicality for real-world applications without the need for extensive tuning.

\noindent\textbf{Computational complexity analysis.} Table \ref{complex} shows the complexity of DiSa in comparison with existing prompt learning approaches, including
CoOp, CoCoOp, IVLP, and PromptSRC, for both training and inference phases. The comparison includes GFLOPs, training time, and throughput (FPS). Notably, DiSa incurs only a 0.11x increase in training GFLOPs compared to the baseline IVLP, while maintaining identical GFLOPs and throughput during inference. In addition, DiSa consistently outperforms these methods in terms of accuracy, highlighting its ability to balance computational efficiency with strong performance.

\begin{table}[t] \setlength\tabcolsep{7.2pt}
\centering
\small
\caption{Impact of $\lambda$ on DiSa's average performance across 11 datasets.}
\begin{tabular}{c | c c c c c c}
\hline \hline
\textbf{$\lambda$} & 0 & 1 & 4 & 8 & 12 & 16 \\
\hline
HM        & 80.21 & 80.41 & 80.62 & 80.90 & \textbf{81.11} & 80.98 \\
\hline \hline
\end{tabular}
\label{lambda}
\end{table}

\begin{table}[t]\setlength\tabcolsep{8.1pt}
\centering
\small
\caption{Ablation on prompt depth for different benchmarks.}
\begin{tabular}{c|c|c}
\hline \hline
 \multirow{2}{*}{\textbf{Layer Depth}} & \textbf{Avg.} & \textbf{HM} \\
\cline{2-3}
{} & Domain Generalization & Base-to-Novel \\
\hline
layer 1 & 60.82 & 77.93\\
layers 1-3 & \textbf{61.43}& 79.08\\
layers 1-6 & 61.30& 80.24\\
layers 1-9 & 60.57& \textbf{81.11}\\
layers 1-12 & 59.14 &81.02\\ 
\hline \hline
\end{tabular}
\label{pdepth}
\end{table}

\begin{table} \setlength\tabcolsep{4.pt}
\small
\caption{Computational cost comparison on the SUN397 dataset. Training times are reported over 10 epochs. DiSa achieves competitive testing efficiency and higher accuracy than all prior methods.}
\begin{tabular}{l|c|c|c|c|c}
\hline \hline
\multirow{2}{*}{\textbf{Method}} & \textbf{GFLOP} & \textbf{GFLOP} & \textbf{Train time} & \multirow{2}{*}{\textbf{FPS}} & \multirow{2}{*}{\textbf{HM}} \\

 & (train) &(test) & (min) &  &  \\
\hline
CoOp \cite{zhou2022learning}      & 162.5 & 162.5 & 10.08 & 1344 & 71.66 \\
CoCoOp \cite{zhou2022conditional}   & 162.5 & 162.5 & 39.53 & 15.08 & 75.83 \\
IVLP \cite{rasheed2023fine}    & 162.8 & 162.8 & 12.01 & 1380 & 77.51 \\
PromptSRC \cite{khattak2023self}  & 179.6 & 162.8 & 13.13 & 1380 &{79.97}\\
\hline
DiSa  & 179.9 & 162.8 & 13.22 & 1380 & \textbf{80.95} \\
\hline \hline
\end{tabular}
\label{complex}
\end{table}

\section{Conclusion}
In this work, we propose a novel directional saliency-aware prompt learning framework designed to enhance the adaptability and generalization of VL models. Our proposed method, DiSa, employs the novel CIR framework to facilitate cooperative learning between prompted and frozen models. The proposed CIR framework leverages a saliency-aware masking approach to ensure the prompted image encoder focuses on semantically critical regions. We also complement our CIR framework with the directional regularization that aligns prompted features with class-wise prototypes in a manner that prioritizes feature orientation over strict proximity. Extensive evaluations across 11 diverse image classification benchmarks demonstrate the superiority of DiSa over state-of-the-art methods, with considerable improvements in generalization to new classes and domains.

\section*{Acknowledgment}
This material is based upon work supported by the National Science Foundation under Grant Numbers CNS-2232048, and CNS-2204445.


\newpage
\bibliographystyle{ACM-Reference-Format}
\bibliography{sample-base}


\begin{thebibliography}{45}


\ifx \showCODEN    \undefined \def \showCODEN     #1{\unskip}     \fi
\ifx \showISBNx    \undefined \def \showISBNx     #1{\unskip}     \fi
\ifx \showISBNxiii \undefined \def \showISBNxiii  #1{\unskip}     \fi
\ifx \showISSN     \undefined \def \showISSN      #1{\unskip}     \fi
\ifx \showLCCN     \undefined \def \showLCCN      #1{\unskip}     \fi
\ifx \shownote     \undefined \def \shownote      #1{#1}          \fi
\ifx \showarticletitle \undefined \def \showarticletitle #1{#1}   \fi
\ifx \showURL      \undefined \def \showURL       {\relax}        \fi
\providecommand\bibfield[2]{#2}
\providecommand\bibinfo[2]{#2}
\providecommand\natexlab[1]{#1}
\providecommand\showeprint[2][]{arXiv:#2}

\bibitem[Augustin et~al\mbox{.}(2020)]%
        {augustin2020adversarial}
\bibfield{author}{\bibinfo{person}{Maximilian Augustin}, \bibinfo{person}{Alexander Meinke}, {and} \bibinfo{person}{Matthias Hein}.} \bibinfo{year}{2020}\natexlab{}.
\newblock \showarticletitle{Adversarial robustness on in-and out-distribution improves explainability}. In \bibinfo{booktitle}{\emph{European Conference on Computer Vision}}. \bibinfo{pages}{228--245}.
\newblock


\bibitem[Bossard et~al\mbox{.}(2014)]%
        {bossard2014food}
\bibfield{author}{\bibinfo{person}{Lukas Bossard}, \bibinfo{person}{Matthieu Guillaumin}, {and} \bibinfo{person}{Luc Van~Gool}.} \bibinfo{year}{2014}\natexlab{}.
\newblock \showarticletitle{Food-101--mining discriminative components with random forests}. In \bibinfo{booktitle}{\emph{European Conference on Computer Vision}}. \bibinfo{pages}{446--461}.
\newblock


\bibitem[Bulat and Tzimiropoulos(2023)]%
        {bulat2023lasp}
\bibfield{author}{\bibinfo{person}{Adrian Bulat} {and} \bibinfo{person}{Georgios Tzimiropoulos}.} \bibinfo{year}{2023}\natexlab{}.
\newblock \showarticletitle{{LASP}: Text-to-text optimization for language-aware soft prompting of vision \& language models}. In \bibinfo{booktitle}{\emph{Proceedings of the IEEE/CVF Conference on Computer Vision and Pattern Recognition}}. \bibinfo{pages}{23232--23241}.
\newblock


\bibitem[Chen et~al\mbox{.}(2022)]%
        {chen2022plot}
\bibfield{author}{\bibinfo{person}{Guangyi Chen}, \bibinfo{person}{Weiran Yao}, \bibinfo{person}{Xiangchen Song}, \bibinfo{person}{Xinyue Li}, \bibinfo{person}{Yongming Rao}, {and} \bibinfo{person}{Kun Zhang}.} \bibinfo{year}{2022}\natexlab{}.
\newblock \showarticletitle{{PLOT}: Prompt learning with optimal transport for vision-language models}.
\newblock \bibinfo{journal}{\emph{arXiv preprint arXiv:2210.01253}} (\bibinfo{year}{2022}).
\newblock


\bibitem[Choi et~al\mbox{.}(2022)]%
        {choi2022tokenmixup}
\bibfield{author}{\bibinfo{person}{Hyeong~Kyu Choi}, \bibinfo{person}{Joonmyung Choi}, {and} \bibinfo{person}{Hyunwoo~J Kim}.} \bibinfo{year}{2022}\natexlab{}.
\newblock \showarticletitle{{TokenMixup}: Efficient attention-guided token-level data augmentation for transformers}.
\newblock \bibinfo{journal}{\emph{Advances in Neural Information Processing Systems}}  \bibinfo{volume}{35} (\bibinfo{year}{2022}), \bibinfo{pages}{14224--14235}.
\newblock


\bibitem[Cimpoi et~al\mbox{.}(2014)]%
        {cimpoi2014describing}
\bibfield{author}{\bibinfo{person}{Mircea Cimpoi}, \bibinfo{person}{Subhransu Maji}, \bibinfo{person}{Iasonas Kokkinos}, \bibinfo{person}{Sammy Mohamed}, {and} \bibinfo{person}{Andrea Vedaldi}.} \bibinfo{year}{2014}\natexlab{}.
\newblock \showarticletitle{Describing textures in the wild}. In \bibinfo{booktitle}{\emph{Proceedings of the IEEE/CVF Conference on Computer Vision and Pattern Recognition}}. \bibinfo{pages}{3606--3613}.
\newblock


\bibitem[Deng et~al\mbox{.}(2009)]%
        {deng2009imagenet}
\bibfield{author}{\bibinfo{person}{Jia Deng}, \bibinfo{person}{Wei Dong}, \bibinfo{person}{Richard Socher}, \bibinfo{person}{Li-Jia Li}, \bibinfo{person}{Kai Li}, {and} \bibinfo{person}{Li Fei-Fei}.} \bibinfo{year}{2009}\natexlab{}.
\newblock \showarticletitle{{ImageNet}: A large-scale hierarchical image database}. In \bibinfo{booktitle}{\emph{Proceedings of the IEEE/CVF Conference on Computer Vision and Pattern Recognition}}. Ieee, \bibinfo{pages}{248--255}.
\newblock


\bibitem[Derakhshani et~al\mbox{.}(2023)]%
        {derakhshani2023bayesian}
\bibfield{author}{\bibinfo{person}{Mohammad~Mahdi Derakhshani}, \bibinfo{person}{Enrique Sanchez}, \bibinfo{person}{Adrian Bulat}, \bibinfo{person}{Victor G~Turrisi da Costa}, \bibinfo{person}{Cees~GM Snoek}, \bibinfo{person}{Georgios Tzimiropoulos}, {and} \bibinfo{person}{Brais Martinez}.} \bibinfo{year}{2023}\natexlab{}.
\newblock \showarticletitle{Bayesian prompt learning for image-language model generalization}. In \bibinfo{booktitle}{\emph{Proceedings of the IEEE/CVF International Conference on Computer Vision}}. \bibinfo{pages}{15237--15246}.
\newblock


\bibitem[Ding et~al\mbox{.}(2022)]%
        {ding2022decoupling}
\bibfield{author}{\bibinfo{person}{Jian Ding}, \bibinfo{person}{Nan Xue}, \bibinfo{person}{Gui-Song Xia}, {and} \bibinfo{person}{Dengxin Dai}.} \bibinfo{year}{2022}\natexlab{}.
\newblock \showarticletitle{Decoupling zero-shot semantic segmentation}. In \bibinfo{booktitle}{\emph{Proceedings of the IEEE/CVF Conference on Computer Vision and Pattern Recognition}}. \bibinfo{pages}{11583--11592}.
\newblock


\bibitem[Fei-Fei et~al\mbox{.}(2004)]%
        {fei2004learning}
\bibfield{author}{\bibinfo{person}{Li Fei-Fei}, \bibinfo{person}{Rob Fergus}, {and} \bibinfo{person}{Pietro Perona}.} \bibinfo{year}{2004}\natexlab{}.
\newblock \showarticletitle{Learning generative visual models from few training examples: An incremental bayesian approach tested on 101 object categories}. In \bibinfo{booktitle}{\emph{Proceedings of the IEEE/CVF Conference on Computer Vision and Pattern Recognition Workshops}}. \bibinfo{pages}{178--178}.
\newblock


\bibitem[Feng et~al\mbox{.}(2022)]%
        {feng2022promptdet}
\bibfield{author}{\bibinfo{person}{Chengjian Feng}, \bibinfo{person}{Yujie Zhong}, \bibinfo{person}{Zequn Jie}, \bibinfo{person}{Xiangxiang Chu}, \bibinfo{person}{Haibing Ren}, \bibinfo{person}{Xiaolin Wei}, \bibinfo{person}{Weidi Xie}, {and} \bibinfo{person}{Lin Ma}.} \bibinfo{year}{2022}\natexlab{}.
\newblock \showarticletitle{{PromptDet}: Towards open-vocabulary detection using uncurated images}. In \bibinfo{booktitle}{\emph{European Conference on Computer Vision}}. \bibinfo{pages}{701--717}.
\newblock


\bibitem[Gao et~al\mbox{.}(2024)]%
        {gao2024clip}
\bibfield{author}{\bibinfo{person}{Peng Gao}, \bibinfo{person}{Shijie Geng}, \bibinfo{person}{Renrui Zhang}, \bibinfo{person}{Teli Ma}, \bibinfo{person}{Rongyao Fang}, \bibinfo{person}{Yongfeng Zhang}, \bibinfo{person}{Hongsheng Li}, {and} \bibinfo{person}{Yu Qiao}.} \bibinfo{year}{2024}\natexlab{}.
\newblock \showarticletitle{{CLIP-Adapter}: Better vision-language models with feature adapters}.
\newblock \bibinfo{journal}{\emph{International Journal of Computer Vision}} \bibinfo{volume}{132}, \bibinfo{number}{2} (\bibinfo{year}{2024}), \bibinfo{pages}{581--595}.
\newblock


\bibitem[He et~al\mbox{.}(2023)]%
        {he2023clip}
\bibfield{author}{\bibinfo{person}{Wenbin He}, \bibinfo{person}{Suphanut Jamonnak}, \bibinfo{person}{Liang Gou}, {and} \bibinfo{person}{Liu Ren}.} \bibinfo{year}{2023}\natexlab{}.
\newblock \showarticletitle{{Clip-S4}: Language-guided self-supervised semantic segmentation}. In \bibinfo{booktitle}{\emph{Proceedings of the IEEE/CVF Conference on Computer Vision and Pattern Recognition}}. \bibinfo{pages}{11207--11216}.
\newblock


\bibitem[Helber et~al\mbox{.}(2019)]%
        {helber2019eurosat}
\bibfield{author}{\bibinfo{person}{Patrick Helber}, \bibinfo{person}{Benjamin Bischke}, \bibinfo{person}{Andreas Dengel}, {and} \bibinfo{person}{Damian Borth}.} \bibinfo{year}{2019}\natexlab{}.
\newblock \showarticletitle{Eurosat: A novel dataset and deep learning benchmark for land use and land cover classification}.
\newblock \bibinfo{journal}{\emph{IEEE Journal of Selected Topics in Applied Earth Observations and Remote Sensing}} \bibinfo{volume}{12}, \bibinfo{number}{7} (\bibinfo{year}{2019}), \bibinfo{pages}{2217--2226}.
\newblock


\bibitem[Hendrycks et~al\mbox{.}(2021a)]%
        {hendrycks2021many}
\bibfield{author}{\bibinfo{person}{Dan Hendrycks}, \bibinfo{person}{Steven Basart}, \bibinfo{person}{Norman Mu}, \bibinfo{person}{Saurav Kadavath}, \bibinfo{person}{Frank Wang}, \bibinfo{person}{Evan Dorundo}, \bibinfo{person}{Rahul Desai}, \bibinfo{person}{Tyler Zhu}, \bibinfo{person}{Samyak Parajuli}, \bibinfo{person}{Mike Guo}, {et~al\mbox{.}}} \bibinfo{year}{2021}\natexlab{a}.
\newblock \showarticletitle{The many faces of robustness: A critical analysis of out-of-distribution generalization}. In \bibinfo{booktitle}{\emph{Proceedings of the IEEE/CVF International Conference on Computer Vision}}. \bibinfo{pages}{8340--8349}.
\newblock


\bibitem[Hendrycks et~al\mbox{.}(2021b)]%
        {hendrycks2021natural}
\bibfield{author}{\bibinfo{person}{Dan Hendrycks}, \bibinfo{person}{Kevin Zhao}, \bibinfo{person}{Steven Basart}, \bibinfo{person}{Jacob Steinhardt}, {and} \bibinfo{person}{Dawn Song}.} \bibinfo{year}{2021}\natexlab{b}.
\newblock \showarticletitle{Natural adversarial examples}. In \bibinfo{booktitle}{\emph{Proceedings of the IEEE/CVF Conference on Computer Vision and Pattern Recognition}}. \bibinfo{pages}{15262--15271}.
\newblock


\bibitem[Ilharco et~al\mbox{.}(2022)]%
        {ilharco2022patching}
\bibfield{author}{\bibinfo{person}{Gabriel Ilharco}, \bibinfo{person}{Mitchell Wortsman}, \bibinfo{person}{Samir~Yitzhak Gadre}, \bibinfo{person}{Shuran Song}, \bibinfo{person}{Hannaneh Hajishirzi}, \bibinfo{person}{Simon Kornblith}, \bibinfo{person}{Ali Farhadi}, {and} \bibinfo{person}{Ludwig Schmidt}.} \bibinfo{year}{2022}\natexlab{}.
\newblock \showarticletitle{Patching open-vocabulary models by interpolating weights}.
\newblock \bibinfo{journal}{\emph{Advances in Neural Information Processing Systems}}  \bibinfo{volume}{35} (\bibinfo{year}{2022}), \bibinfo{pages}{29262--29277}.
\newblock


\bibitem[Jia et~al\mbox{.}(2021)]%
        {jia2021scaling}
\bibfield{author}{\bibinfo{person}{Chao Jia}, \bibinfo{person}{Yinfei Yang}, \bibinfo{person}{Ye Xia}, \bibinfo{person}{Yi-Ting Chen}, \bibinfo{person}{Zarana Parekh}, \bibinfo{person}{Hieu Pham}, \bibinfo{person}{Quoc Le}, \bibinfo{person}{Yun-Hsuan Sung}, \bibinfo{person}{Zhen Li}, {and} \bibinfo{person}{Tom Duerig}.} \bibinfo{year}{2021}\natexlab{}.
\newblock \showarticletitle{Scaling up visual and vision-language representation learning with noisy text supervision}. In \bibinfo{booktitle}{\emph{International Conference on Machine Learning}}. PMLR, \bibinfo{pages}{4904--4916}.
\newblock


\bibitem[Khattak et~al\mbox{.}(2023a)]%
        {khattak2023maple}
\bibfield{author}{\bibinfo{person}{Muhammad~Uzair Khattak}, \bibinfo{person}{Hanoona Rasheed}, \bibinfo{person}{Muhammad Maaz}, \bibinfo{person}{Salman Khan}, {and} \bibinfo{person}{Fahad~Shahbaz Khan}.} \bibinfo{year}{2023}\natexlab{a}.
\newblock \showarticletitle{Maple: Multi-modal prompt learning}. In \bibinfo{booktitle}{\emph{Proceedings of the IEEE/CVF Conference on Computer Vision and Pattern Recognition}}. \bibinfo{pages}{19113--19122}.
\newblock


\bibitem[Khattak et~al\mbox{.}(2023b)]%
        {khattak2023self}
\bibfield{author}{\bibinfo{person}{Muhammad~Uzair Khattak}, \bibinfo{person}{Syed~Talal Wasim}, \bibinfo{person}{Muzammal Naseer}, \bibinfo{person}{Salman Khan}, \bibinfo{person}{Ming-Hsuan Yang}, {and} \bibinfo{person}{Fahad~Shahbaz Khan}.} \bibinfo{year}{2023}\natexlab{b}.
\newblock \showarticletitle{Self-regulating prompts: Foundational model adaptation without forgetting}. In \bibinfo{booktitle}{\emph{Proceedings of the IEEE/CVF International Conference on Computer Vision}}. \bibinfo{pages}{15190--15200}.
\newblock


\bibitem[Kim et~al\mbox{.}(2021)]%
        {kim2021adapt}
\bibfield{author}{\bibinfo{person}{Konwoo Kim}, \bibinfo{person}{Michael Laskin}, \bibinfo{person}{Igor Mordatch}, {and} \bibinfo{person}{Deepak Pathak}.} \bibinfo{year}{2021}\natexlab{}.
\newblock \showarticletitle{How to adapt your large-scale vision-and-language model}.
\newblock  (\bibinfo{year}{2021}).
\newblock


\bibitem[Krause et~al\mbox{.}(2013)]%
        {krause20133d}
\bibfield{author}{\bibinfo{person}{Jonathan Krause}, \bibinfo{person}{Michael Stark}, \bibinfo{person}{Jia Deng}, {and} \bibinfo{person}{Li Fei-Fei}.} \bibinfo{year}{2013}\natexlab{}.
\newblock \showarticletitle{{3D} object representations for fine-grained categorization}. In \bibinfo{booktitle}{\emph{Proceedings of the IEEE/CVF International Conference on Computer Vision Workshops}}. \bibinfo{pages}{554--561}.
\newblock


\bibitem[Loshchilov and Hutter(2017)]%
        {loshchilov2017decoupled}
\bibfield{author}{\bibinfo{person}{Ilya Loshchilov} {and} \bibinfo{person}{Frank Hutter}.} \bibinfo{year}{2017}\natexlab{}.
\newblock \showarticletitle{Decoupled weight decay regularization}.
\newblock \bibinfo{journal}{\emph{arXiv preprint arXiv:1711.05101}} (\bibinfo{year}{2017}).
\newblock


\bibitem[Lu et~al\mbox{.}(2022)]%
        {lu2022prompt}
\bibfield{author}{\bibinfo{person}{Yuning Lu}, \bibinfo{person}{Jianzhuang Liu}, \bibinfo{person}{Yonggang Zhang}, \bibinfo{person}{Yajing Liu}, {and} \bibinfo{person}{Xinmei Tian}.} \bibinfo{year}{2022}\natexlab{}.
\newblock \showarticletitle{Prompt distribution learning}. In \bibinfo{booktitle}{\emph{Proceedings of the IEEE/CVF Conference on Computer Vision and Pattern Recognition}}. \bibinfo{pages}{5206--5215}.
\newblock


\bibitem[Maji et~al\mbox{.}(2013)]%
        {maji2013fine}
\bibfield{author}{\bibinfo{person}{Subhransu Maji}, \bibinfo{person}{Esa Rahtu}, \bibinfo{person}{Juho Kannala}, \bibinfo{person}{Matthew Blaschko}, {and} \bibinfo{person}{Andrea Vedaldi}.} \bibinfo{year}{2013}\natexlab{}.
\newblock \showarticletitle{Fine-grained visual classification of aircraft}.
\newblock \bibinfo{journal}{\emph{arXiv preprint arXiv:1306.5151}} (\bibinfo{year}{2013}).
\newblock


\bibitem[Naeem et~al\mbox{.}(2022)]%
        {naeem2022i2dformer}
\bibfield{author}{\bibinfo{person}{Muhammad~Ferjad Naeem}, \bibinfo{person}{Yongqin Xian}, \bibinfo{person}{Luc~V Gool}, {and} \bibinfo{person}{Federico Tombari}.} \bibinfo{year}{2022}\natexlab{}.
\newblock \showarticletitle{{I2DFormer}: Learning image to document attention for zero-shot image classification}.
\newblock \bibinfo{journal}{\emph{Advances in Neural Information Processing Systems}}  \bibinfo{volume}{35} (\bibinfo{year}{2022}), \bibinfo{pages}{12283--12294}.
\newblock


\bibitem[Nilsback and Zisserman(2008)]%
        {nilsback2008automated}
\bibfield{author}{\bibinfo{person}{Maria-Elena Nilsback} {and} \bibinfo{person}{Andrew Zisserman}.} \bibinfo{year}{2008}\natexlab{}.
\newblock \showarticletitle{Automated flower classification over a large number of classes}. In \bibinfo{booktitle}{\emph{Sixth Indian Conference on Computer Vision, Graphics \& Image Processing}}. \bibinfo{pages}{722--729}.
\newblock


\bibitem[Parkhi et~al\mbox{.}(2012)]%
        {parkhi2012cats}
\bibfield{author}{\bibinfo{person}{Omkar~M Parkhi}, \bibinfo{person}{Andrea Vedaldi}, \bibinfo{person}{Andrew Zisserman}, {and} \bibinfo{person}{CV Jawahar}.} \bibinfo{year}{2012}\natexlab{}.
\newblock \showarticletitle{Cats and dogs}. In \bibinfo{booktitle}{\emph{Proceedings of the IEEE/CVF Conference on Computer Vision and Pattern Recognition}}. \bibinfo{pages}{3498--3505}.
\newblock


\bibitem[Radford et~al\mbox{.}(2021)]%
        {radford2021learning}
\bibfield{author}{\bibinfo{person}{Alec Radford}, \bibinfo{person}{Jong~Wook Kim}, \bibinfo{person}{Chris Hallacy}, \bibinfo{person}{Aditya Ramesh}, \bibinfo{person}{Gabriel Goh}, \bibinfo{person}{Sandhini Agarwal}, \bibinfo{person}{Girish Sastry}, \bibinfo{person}{Amanda Askell}, \bibinfo{person}{Pamela Mishkin}, \bibinfo{person}{Jack Clark}, {et~al\mbox{.}}} \bibinfo{year}{2021}\natexlab{}.
\newblock \showarticletitle{Learning transferable visual models from natural language supervision}. In \bibinfo{booktitle}{\emph{International Conference on Machine Learning}}. PMLR, \bibinfo{pages}{8748--8763}.
\newblock


\bibitem[Rao et~al\mbox{.}(2022)]%
        {rao2022denseclip}
\bibfield{author}{\bibinfo{person}{Yongming Rao}, \bibinfo{person}{Wenliang Zhao}, \bibinfo{person}{Guangyi Chen}, \bibinfo{person}{Yansong Tang}, \bibinfo{person}{Zheng Zhu}, \bibinfo{person}{Guan Huang}, \bibinfo{person}{Jie Zhou}, {and} \bibinfo{person}{Jiwen Lu}.} \bibinfo{year}{2022}\natexlab{}.
\newblock \showarticletitle{{DenseCLIP}: Language-guided dense prediction with context-aware prompting}. In \bibinfo{booktitle}{\emph{Proceedings of the IEEE/CVF Conference on Computer Vision and Pattern Recognition}}. \bibinfo{pages}{18082--18091}.
\newblock


\bibitem[Rasheed et~al\mbox{.}(2023)]%
        {rasheed2023fine}
\bibfield{author}{\bibinfo{person}{Hanoona Rasheed}, \bibinfo{person}{Muhammad~Uzair Khattak}, \bibinfo{person}{Muhammad Maaz}, \bibinfo{person}{Salman Khan}, {and} \bibinfo{person}{Fahad~Shahbaz Khan}.} \bibinfo{year}{2023}\natexlab{}.
\newblock \showarticletitle{Fine-tuned {CLIP} models are efficient video learners}. In \bibinfo{booktitle}{\emph{Proceedings of the IEEE/CVF Conference on Computer Vision and Pattern Recognition}}. \bibinfo{pages}{6545--6554}.
\newblock


\bibitem[Recht et~al\mbox{.}(2019)]%
        {recht2019imagenet}
\bibfield{author}{\bibinfo{person}{Benjamin Recht}, \bibinfo{person}{Rebecca Roelofs}, \bibinfo{person}{Ludwig Schmidt}, {and} \bibinfo{person}{Vaishaal Shankar}.} \bibinfo{year}{2019}\natexlab{}.
\newblock \showarticletitle{Do {ImageNet} classifiers generalize to {ImageNet}?}. In \bibinfo{booktitle}{\emph{International Conference on Machine Learning}}. PMLR, \bibinfo{pages}{5389--5400}.
\newblock


\bibitem[Roy and Etemad(2023)]%
        {roy2023consistency}
\bibfield{author}{\bibinfo{person}{Shuvendu Roy} {and} \bibinfo{person}{Ali Etemad}.} \bibinfo{year}{2023}\natexlab{}.
\newblock \showarticletitle{Consistency-guided prompt learning for vision-language models}.
\newblock \bibinfo{journal}{\emph{arXiv preprint arXiv:2306.01195}} (\bibinfo{year}{2023}).
\newblock


\bibitem[Soomro et~al\mbox{.}(2012)]%
        {soomro2012ucf101}
\bibfield{author}{\bibinfo{person}{Khurram Soomro}, \bibinfo{person}{Amir~Roshan Zamir}, {and} \bibinfo{person}{Mubarak Shah}.} \bibinfo{year}{2012}\natexlab{}.
\newblock \showarticletitle{UCF101: A dataset of 101 human actions classes from videos in the wild}.
\newblock \bibinfo{journal}{\emph{arXiv preprint arXiv:1212.0402}} (\bibinfo{year}{2012}).
\newblock


\bibitem[Srivastava et~al\mbox{.}(2014)]%
        {srivastava2014dropout}
\bibfield{author}{\bibinfo{person}{Nitish Srivastava}, \bibinfo{person}{Geoffrey Hinton}, \bibinfo{person}{Alex Krizhevsky}, \bibinfo{person}{Ilya Sutskever}, {and} \bibinfo{person}{Ruslan Salakhutdinov}.} \bibinfo{year}{2014}\natexlab{}.
\newblock \showarticletitle{Dropout: a simple way to prevent neural networks from overfitting}.
\newblock \bibinfo{journal}{\emph{The Journal of Machine Learning Research}} \bibinfo{volume}{15}, \bibinfo{number}{1} (\bibinfo{year}{2014}), \bibinfo{pages}{1929--1958}.
\newblock


\bibitem[Tsimpoukelli et~al\mbox{.}(2021)]%
        {tsimpoukelli2021multimodal}
\bibfield{author}{\bibinfo{person}{Maria Tsimpoukelli}, \bibinfo{person}{Jacob~L Menick}, \bibinfo{person}{Serkan Cabi}, \bibinfo{person}{SM Eslami}, \bibinfo{person}{Oriol Vinyals}, {and} \bibinfo{person}{Felix Hill}.} \bibinfo{year}{2021}\natexlab{}.
\newblock \showarticletitle{Multimodal few-shot learning with frozen language models}.
\newblock \bibinfo{journal}{\emph{Advances in Neural Information Processing Systems}}  \bibinfo{volume}{34} (\bibinfo{year}{2021}), \bibinfo{pages}{200--212}.
\newblock


\bibitem[Wang et~al\mbox{.}(2019)]%
        {wang2019learning}
\bibfield{author}{\bibinfo{person}{Haohan Wang}, \bibinfo{person}{Songwei Ge}, \bibinfo{person}{Zachary Lipton}, {and} \bibinfo{person}{Eric~P Xing}.} \bibinfo{year}{2019}\natexlab{}.
\newblock \showarticletitle{Learning robust global representations by penalizing local predictive power}.
\newblock \bibinfo{journal}{\emph{Advances in Neural Information Processing Systems}}  \bibinfo{volume}{32} (\bibinfo{year}{2019}).
\newblock


\bibitem[Xiao et~al\mbox{.}(2010)]%
        {xiao2010sun}
\bibfield{author}{\bibinfo{person}{Jianxiong Xiao}, \bibinfo{person}{James Hays}, \bibinfo{person}{Krista~A Ehinger}, \bibinfo{person}{Aude Oliva}, {and} \bibinfo{person}{Antonio Torralba}.} \bibinfo{year}{2010}\natexlab{}.
\newblock \showarticletitle{Sun database: Large-scale scene recognition from abbey to zoo}. In \bibinfo{booktitle}{\emph{Proceedings of the IEEE/CVF Conference on Computer Vision and Pattern Recognition}}. \bibinfo{pages}{3485--3492}.
\newblock


\bibitem[Yang et~al\mbox{.}(2024)]%
        {yang2024mma}
\bibfield{author}{\bibinfo{person}{Lingxiao Yang}, \bibinfo{person}{Ru-Yuan Zhang}, \bibinfo{person}{Yanchen Wang}, {and} \bibinfo{person}{Xiaohua Xie}.} \bibinfo{year}{2024}\natexlab{}.
\newblock \showarticletitle{{MMA}: Multi-Modal Adapter for Vision-Language Models}. In \bibinfo{booktitle}{\emph{Proceedings of the IEEE/CVF Conference on Computer Vision and Pattern Recognition}}. \bibinfo{pages}{23826--23837}.
\newblock


\bibitem[Yang et~al\mbox{.}(2023)]%
        {yang2023towards}
\bibfield{author}{\bibinfo{person}{Yongjin Yang}, \bibinfo{person}{Jongwoo Ko}, {and} \bibinfo{person}{Se-Young Yun}.} \bibinfo{year}{2023}\natexlab{}.
\newblock \showarticletitle{Towards Difficulty-Agnostic Efficient Transfer Learning for Vision-Language Models}.
\newblock \bibinfo{journal}{\emph{arXiv preprint arXiv:2311.15569}} (\bibinfo{year}{2023}).
\newblock


\bibitem[Yao et~al\mbox{.}(2024)]%
        {yao2024tcp}
\bibfield{author}{\bibinfo{person}{Hantao Yao}, \bibinfo{person}{Rui Zhang}, {and} \bibinfo{person}{Changsheng Xu}.} \bibinfo{year}{2024}\natexlab{}.
\newblock \showarticletitle{Tcp: Textual-based class-aware prompt tuning for visual-language model}. In \bibinfo{booktitle}{\emph{Proceedings of the IEEE/CVF Conference on Computer Vision and Pattern Recognition}}. \bibinfo{pages}{23438--23448}.
\newblock


\bibitem[Yao et~al\mbox{.}(2021)]%
        {yao2021filip}
\bibfield{author}{\bibinfo{person}{Lewei Yao}, \bibinfo{person}{Runhui Huang}, \bibinfo{person}{Lu Hou}, \bibinfo{person}{Guansong Lu}, \bibinfo{person}{Minzhe Niu}, \bibinfo{person}{Hang Xu}, \bibinfo{person}{Xiaodan Liang}, \bibinfo{person}{Zhenguo Li}, \bibinfo{person}{Xin Jiang}, {and} \bibinfo{person}{Chunjing Xu}.} \bibinfo{year}{2021}\natexlab{}.
\newblock \showarticletitle{Filip: Fine-grained interactive language-image pre-training}.
\newblock \bibinfo{journal}{\emph{arXiv preprint arXiv:2111.07783}} (\bibinfo{year}{2021}).
\newblock


\bibitem[Zheng et~al\mbox{.}(2023)]%
        {zheng2023localization}
\bibfield{author}{\bibinfo{person}{Zhaohui Zheng}, \bibinfo{person}{Rongguang Ye}, \bibinfo{person}{Qibin Hou}, \bibinfo{person}{Dongwei Ren}, \bibinfo{person}{Ping Wang}, \bibinfo{person}{Wangmeng Zuo}, {and} \bibinfo{person}{Ming-Ming Cheng}.} \bibinfo{year}{2023}\natexlab{}.
\newblock \showarticletitle{Localization distillation for object detection}.
\newblock \bibinfo{journal}{\emph{IEEE Transactions on Pattern Analysis and Machine Intelligence}} \bibinfo{volume}{45}, \bibinfo{number}{8} (\bibinfo{year}{2023}), \bibinfo{pages}{10070--10083}.
\newblock


\bibitem[Zhou et~al\mbox{.}(2022a)]%
        {zhou2022conditional}
\bibfield{author}{\bibinfo{person}{Kaiyang Zhou}, \bibinfo{person}{Jingkang Yang}, \bibinfo{person}{Chen~Change Loy}, {and} \bibinfo{person}{Ziwei Liu}.} \bibinfo{year}{2022}\natexlab{a}.
\newblock \showarticletitle{Conditional prompt learning for vision-language models}. In \bibinfo{booktitle}{\emph{Proceedings of the IEEE/CVF Conference on Computer Vision and Pattern Recognition}}. \bibinfo{pages}{16816--16825}.
\newblock


\bibitem[Zhou et~al\mbox{.}(2022b)]%
        {zhou2022learning}
\bibfield{author}{\bibinfo{person}{Kaiyang Zhou}, \bibinfo{person}{Jingkang Yang}, \bibinfo{person}{Chen~Change Loy}, {and} \bibinfo{person}{Ziwei Liu}.} \bibinfo{year}{2022}\natexlab{b}.
\newblock \showarticletitle{Learning to prompt for vision-language models}.
\newblock \bibinfo{journal}{\emph{International Journal of Computer Vision}} \bibinfo{volume}{130}, \bibinfo{number}{9} (\bibinfo{year}{2022}), \bibinfo{pages}{2337--2348}.
\newblock


\end{thebibliography}










\end{document}